\documentclass[letterpaper, 10 pt, journal, twoside]{IEEEtran}  

\usepackage{amsmath} 
\usepackage{amssymb}  
\usepackage{mathrsfs}
\usepackage{graphicx} 
\usepackage[caption=false,labelformat=parens]{subfig}
\usepackage{color}
\usepackage{bm}
\usepackage{cite}

\newtheorem{assumption}{\textbf{Assumption}}

\newtheorem{remark}{\textbf{Remark}}

\newcommand{\Real}{\mathbb R}

\newcommand{\col}{\textrm{col}}

\newcommand{\iden}{\mathbb{I}}
\newcommand{\net}{\textrm{net}}
\newcommand{\vect}{\textrm{vec}}
\newcommand{\op}{\textrm{op}}
\newcommand{\des}{\textrm{des}}

\newcommand{\inter}{\textrm{int}}

\title{Layered Control for Cooperative Locomotion of Two Quadrupedal Robots: Centralized and Distributed Approaches}

\author{Jeeseop~Kim$^{1}$, Randall~T.~Fawcett$^{1}$, Vinay~R.~Kamidi$^{1}$, Aaron~D.~Ames$^{2}$, and Kaveh~Akbari Hamed$^{1}$
\thanks{The work of J. Kim and K. Akbari Hamed is supported by the National Science Foundation (NSF) under the Grant 1924617. The work of R. T. Fawcett is supported by the NSF under Grant 2128948. The work of A. D. Ames is supported by the NSF under Grant 1924526.}
\thanks{$^{1}$J. Kim, R. T. Fawcett, V.~R.~Kamidi, and K. Akbari Hamed (Corresponding Author) are with the Department of Mechanical Engineering, Virginia Tech, Blacksburg, VA 24061, USA, {\tt\small \{jeeseop, randallf, vinay28, kavehakbarihamed\}@vt.edu}}%
\thanks{$^{2}$A. D. Ames is with the Department of Mechanical and Civil Engineering, California Institute of Technology, Pasadena, CA 91125, USA, {\tt\small ames@caltech.edu}}%
}

\begin{document}
\maketitle



\begin{abstract}
This paper presents a layered control approach for real-time trajectory planning and control of robust cooperative locomotion by two holonomically constrained quadrupedal robots. A novel interconnected network of reduced-order models, based on the single rigid body (SRB) dynamics, is developed for trajectory planning purposes. At the higher level of the control architecture, two different model predictive control (MPC) algorithms are proposed to address the optimal control problem of the interconnected SRB dynamics: centralized and distributed MPCs. The distributed MPC assumes two local quadratic programs that share their optimal solutions according to a one-step communication delay and an agreement protocol. At the lower level of the control scheme, distributed nonlinear controllers are developed to impose the full-order dynamics to track the prescribed reduced-order trajectories generated by MPCs. The effectiveness of the control approach is verified with extensive numerical simulations and experiments for the robust and cooperative locomotion of two holonomically constrained A1 robots with different payloads on variable terrains and in the presence of disturbances. It is shown that the distributed MPC has a performance similar to that of the centralized MPC, while the computation time is reduced significantly.
\end{abstract}


\vspace{-0.5em}
\section{Introduction}
\label{Sec:Introduction}


\subsection{Motivation and Goal}

\begin{figure}[t!]
\centering
\includegraphics[width=\linewidth]{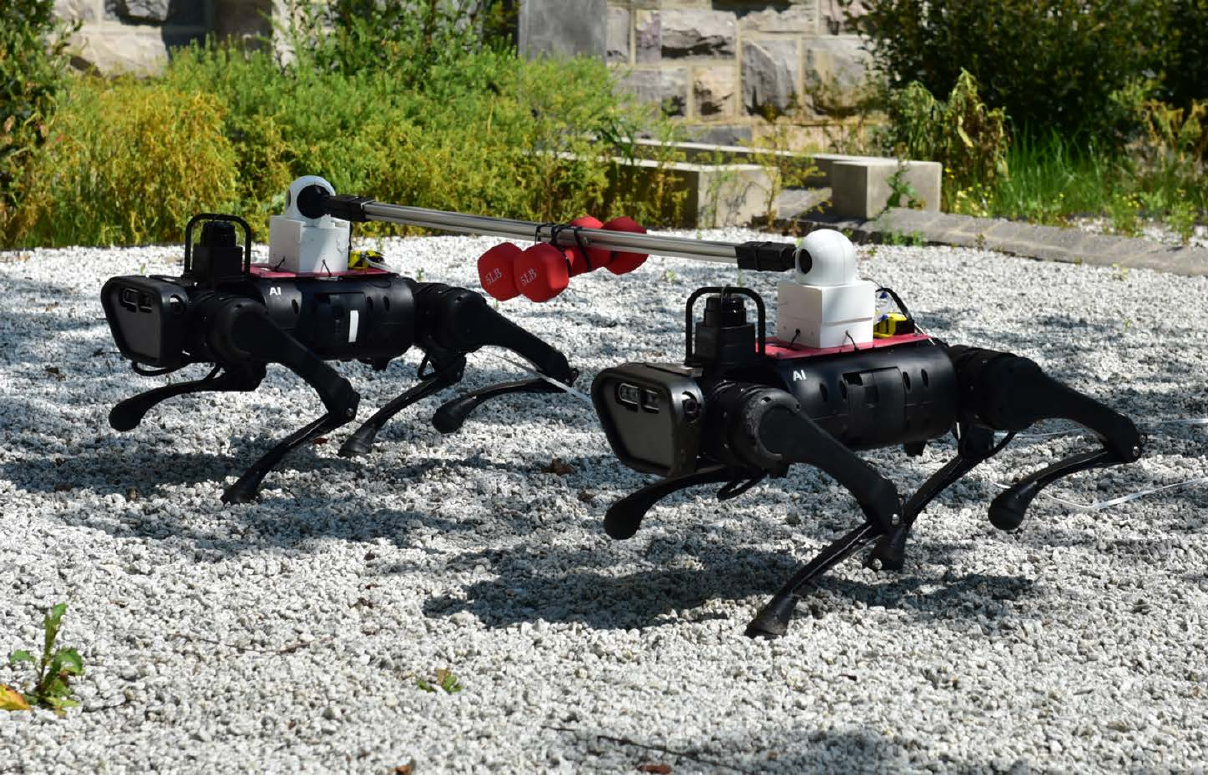}
\vspace{-1.8em}
\caption{Snapshot illustrating holonomically constrained quadrupedal robots locomoting on gravel while carrying a payload of 4.53 (kg).}
\label{fig:2agents_snap}
\vspace{-1.5em}
\end{figure}

Human-centered communities, including factories, offices, and homes, are typically developed for humans who are bipedal walkers capable of stepping over gaps and walking up/down stairs. This motivates the development of collaborative legged robots that can cooperatively work with each other to assist humans in different aspects of their life, such as labor-intensive tasks, construction, manufacturing, and assembly. One of the most challenging and essential problems in deploying collaborative legged robots is \textit{cooperative locomotion} in complex environments, wherein the collaboration between robots is described by holonomic constraints. Cooperative locomotion with holonomic constraints arises in different applications of legged robots, such as cooperative transportation of payloads like social insects \cite{Jeeseop_Hamed_ASME} (see Fig. \ref{fig:2agents_snap}), human-robot locomotion via prosthetic legs and exoskeletons \cite{Grizzle_Decentralized,Hamed_Gregg_IEEE_TAC,Gregg_Toward_Biomimetic_Control_IEEE_CST, zhao2016multicontact}, and human-robot locomotion via guide dog robots \cite{Hamed_GuideDogHuman_IEERAL}. 

\begin{figure*}[t!]
\centering
\includegraphics[width=\linewidth]{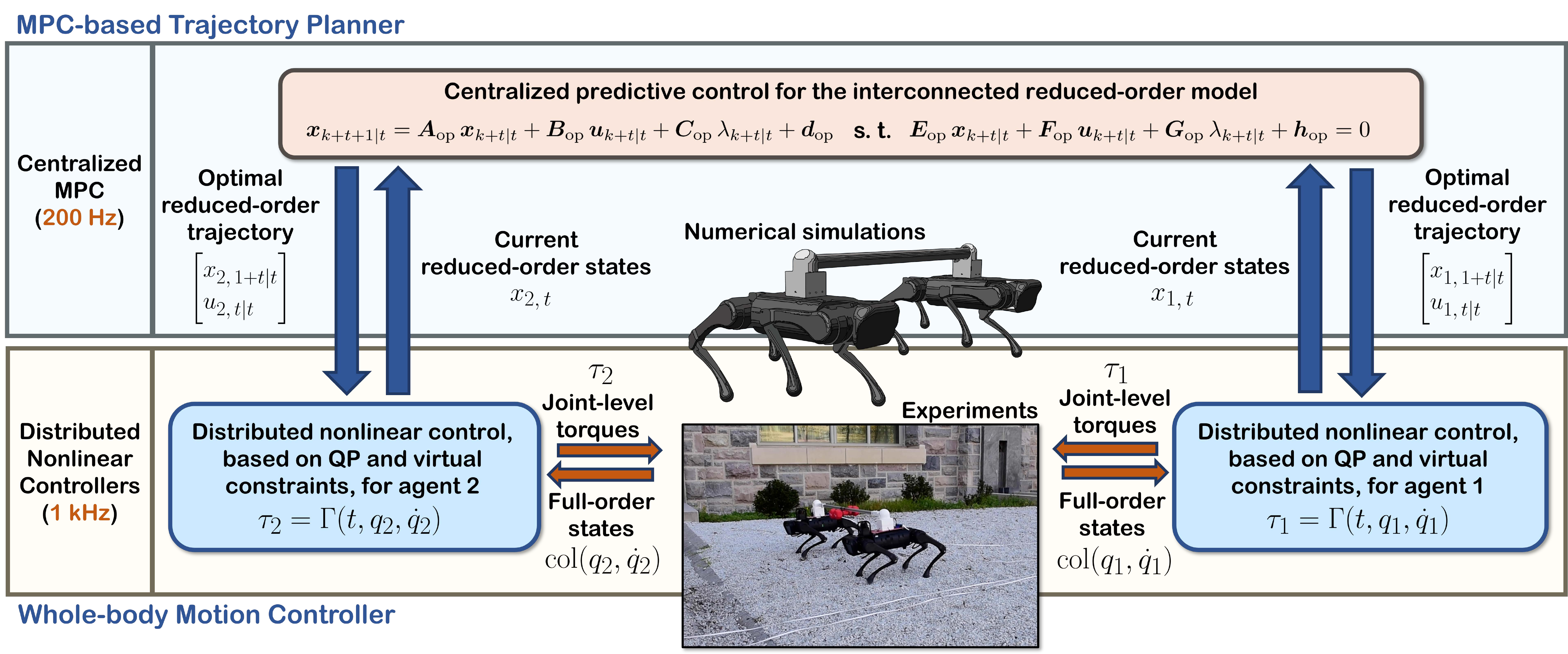}
\vspace{-1.8em}
\caption{Overview of the proposed layered control approach with the centralized MPC algorithm at the high level and distributed nonlinear controllers at the low level for cooperative locomotion.}
\label{fig:overview_centralized}
\vspace{-1.5em}
\end{figure*}

In recent years, important theoretical and technological advances have allowed for the successful control of multi-robot systems (MRSs) \cite{literature_review_MRS_01, literature_review_MRS_02}, including collaborative robotic arms with or without mobility \cite{Murray_Book,      williams1993virtual, culbertson2021decentralized,     erhart2013impedance, alonso2017multi, elwin2022human}, aerial vehicles \cite{chen2020guaranteed,     RSS2013, masone2016cooperative, pereira2017collaborative, li2021cooperative,     mellinger2013cooperative, nguyen2018novel, tagliabue2019robust, wehbeh2020distributed,     caccavale2015cooperative, yang2015hierarchical, lee2018integrated}, and ground vehicles \cite{DP:MT:14, spletzer2001cooperative, pereira2004decentralized, farivarnejad2016decentralized, machado2016multi}. In addition, distributed control algorithms, including distributed receding horizon control approaches, have been developed to address the motion planning of MRSs, see e.g., \cite{Mesbahi_Book, Bullo_Book, DUNBAR_distributedMPC, Ditributed_MPC_Book}. Some recent works also address the control and planning of heterogeneous robot teams, including legged robots \cite{ahmadi2019safe, tranzatto2022cerberus, yang2022collaborative} but without holonomic constraints amongst the agents. However, the capabilities of cooperative legged locomotion have not been fully explored. In particular, collaborating legged robots can be described by \textit{inherently unstable} dynamical systems with \textit{high dimensionality} (i.e., high degrees of freedom (DOFs)), \textit{nonlinear}, and \textit{hybrid} nature, and subject to underactuation and unilateral constraints, as opposed to most of the MRSs where the state-of-the-art algorithms have been deployed \cite{Hamed_Kamidi_Pandala_Ma_Ames_ACC}. This complicates the design of real-time trajectory planning and control approaches, both in centralized and distributed fashions, to guarantee each agent's dynamic and robust stability while addressing the curse of dimensionality and respecting the holonomic and unilateral constraints. 

Reduced-order (i.e., template) models provide low-dimensional realizations of full-order dynamical models of legged robots \cite{Full_Koditschek_Template}. They can be integrated with convex optimization techniques and model predictive control (MPC) approaches to enable gait planning for the existing legged robots. Some popular reduced-order models  include the linear inverted pendulum (LIP) model \cite{kajita19991LIP}, centroidal dynamics \cite{orin2013centroidal}, and single rigid body (SRB) dynamics \cite{KF_mitcheetah3,Wensing_VBL_HJB,Abhishek_Hae-Won_TRO}. These template models have been used for real-time planning of different single-agent bipedal \cite{Leonessa_Pratt_MPC,Ott_MPC,Pratt_LIP, dai2014whole} and quadrupedal robots \cite{Abhishek_Hae-Won_TRO,KF_mitcheetah3,Wensing_VBL_HJB,     Kim_Wensing_Convex_MPC_01,grandia2019frequency,hamed2020quadrupedal,Randy_Hamed_ASME,farshidian2017efficient,pandala2022robust}. In this paper, we aim to answer three \textit{fundamental questions} in the context of cooperative locomotion of legged robots. 1) How do we develop effective and interconnected reduced-order models that describe the cooperative locomotion of dynamic legged robots with holonomic constraints? 2) How do we develop computationally tractable predictive control algorithms in centralized and distributed manners for real-time planning of interconnected reduced-order models? In particular, we aim to examine the implementation of centralized and distributed predictive control algorithms for real-time planning to overcome the limitations caused by the curse of dimensionality in cooperative locomotion. And 3) How do we map optimal reduced-order trajectories to full-order and complex dynamical models of cooperative locomotion? 

\begin{figure*}[t!]
\centering
\includegraphics[width=\linewidth]{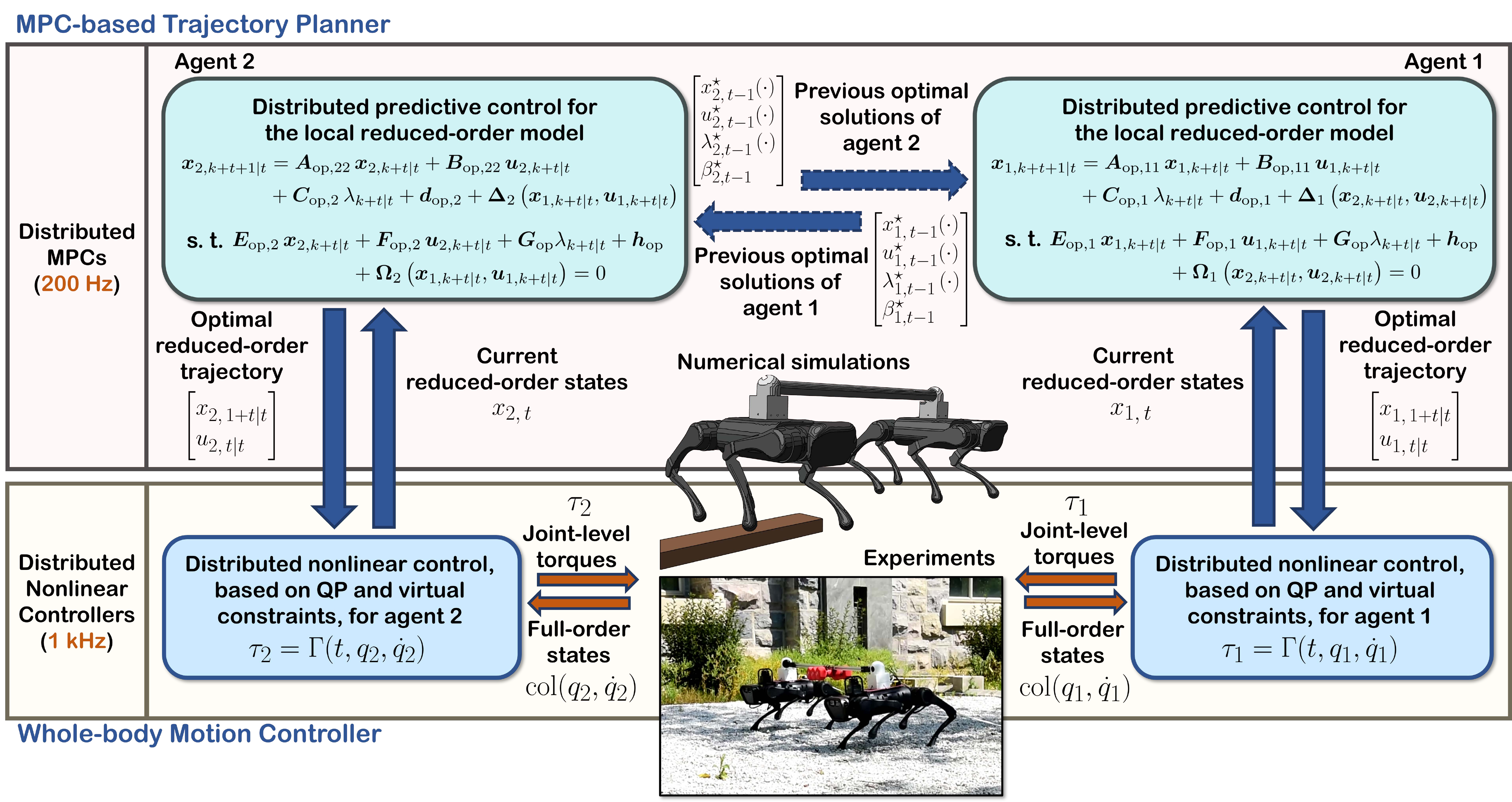}
\vspace{-1.8em}
\caption{Overview of the proposed layered control approach with the distributed MPC algorithms at the high level and distributed nonlinear controllers at the low level for cooperative locomotion.}
\label{fig:overview_distributed}
\vspace{-1.5em}
\end{figure*}

In order to address the above questions, this paper aims to develop mathematical foundations, experimentally implement, and comprehensively study the cooperative locomotion of two holonomically constrained dynamic legged robots.  In particular, the \textit{overarching goal} of this paper is to develop a layered control algorithm for the real-time trajectory planning and control of dynamic cooperative locomotion for two holonomically constrained legged-robotic systems. The higher layer of the proposed algorithm considers an innovative reduced-order model composed of two interconnected SRB dynamics subject to holonomic constraints for the planning problem. The paper develops novel centralized and distributed MPC algorithms for the planning purpose of interconnected SRB dynamics (see Figs. \ref{fig:overview_centralized} and \ref{fig:overview_distributed}). These MPC algorithms address the real-time planning at the higher layer of the control hierarchy subject to the interaction terms and feasibility of the ground reaction forces (GRFs). The optimal reduced-order trajectories and GRFs, generated by the high-level MPCs,  are then mapped to the full-order and complex dynamics via distributed nonlinear controllers at the low level for the whole-body motion control. The low-level nonlinear controllers are developed based on quadratic programming (QP) and input-output (I-O) linearization. The efficacy of the proposed layered control approach is validated via extensive experiments for robustly stable locomotion of two holonomically constrained A1 quadrupedal robots that cooperatively transport unknown payloads on different terrains and in the presence of disturbances (see Fig. \ref{fig:2agents_snap}). A comprehensive numerical analysis of the performance of the proposed centralized and distributed MPC algorithms is finally presented. 

\vspace{-0.5em}
\subsection{Related Work}   
\label{Sec:Related Work and Motivation}

Holonomically constrained MRSs, including fixed-based collaborative robotic arms \cite{williams1993virtual, culbertson2021decentralized}, aerial vehicles with payloads \cite{wehbeh2020distributed, mellinger2013cooperative, nguyen2018novel, tagliabue2019robust}, and ground vehicles \cite{spletzer2001cooperative, pereira2004decentralized, farivarnejad2016decentralized, machado2016multi} have gained significant attention during the last years. Moreover, MRSs augmented with robotic arms have been studied for more complex cooperative tasks  \cite{erhart2013impedance, alonso2017multi, elwin2022human, caccavale2015cooperative, yang2015hierarchical, lee2018integrated}. 
In contrast to the above-mentioned robotic systems, collaborative legged robots are dynamical systems with high dimensionality, unilateral constraints, and hybrid nature that add further complexity to synthesizing planning and control algorithms. In addition, the strong interacting wrenches (forces/torques) between the agents, arising from holonomic constraints, must be carefully addressed to result in a robustly stable planner for cooperative legged locomotion. As a result, collaborative legged locomotion has not been studied to the same degree as other robotic systems. This paper, there, marks the first experimental implementation in this context. 

In the context of legged robots, the trajectory planning and control approaches can be sectioned into two categories: the ones using the full-order models and the others using the reduced-order models.  Hybrid systems theory plays an important role in understanding and analyzing full-order dynamical models of legged locomotion \cite{Grizzle_Asymptotically_Stable_Walking_IEEE_TAC,Jessy_Book,Hurmuzlu_Impact,Morriz_Grizzle_Hybrid_Invariant_Manifolds_IEEE_TAC,Poulakakis_Grizzle_SLIP_IEEE_TAC,Haddad_Hybrid_Book,Goebel_Hybrid_Book,Johnson_Burden_Koditschek,Spong_Passivity_IEEE_RAM}. Advanced nonlinear control algorithms such as hybrid reduction \cite{Ames_HybridReduction_Original_Paper}, controlled symmetries \cite{Spong_Controlled_Symmetries_IEEE_TAC}, transverse linearization \cite{Manchester_Tedrake_LQR_IJRR}, and hybrid zero dynamics (HZD) \cite{Ames_RES_CLF_IEEE_TAC,Westervelt_Grizzle_Koditschek_HZD_IEEE_TRO} address the hybrid nature of full-order locomotion models. The HZD approach regulates some output functions, referred to as virtual constraints, with I-O linearization techniques \cite{Isidori_Book} to coordinate the robot's links within a stride. This method can systematically address underactuation and its effectiveness has been validated for stable locomotion of different bipedal \cite{Cheavallereau_Grizzle_RABBIT,Sreenath_Grizzle_HZD_Running_IJRR,da2019combining,Ames_DURUS_TRO,Martin_Schmiedeler_IJRR} and quadrupedal robots \cite{Randy_Paper_LCSS, Ma_Shishir_Hamed_Ames} as well as powered prosthetic legs \cite{Gregg_Toward_Biomimetic_Control_IEEE_CST, zhao2016multicontact}. The full-order gait planning is typically formulated as a nonlinear programming (NLP) problem that can be addressed with existing NLP tools and direct collocation techniques \cite{Ames_DURUS_TRO,collocation_Posa,collocation_03,collocation_Kelly,patel2019contact}. Although the direct-collocation-based approaches generate optimal trajectories for full-order models of legged robots effectively, they \textit{cannot} address real-time trajectory optimization of cooperative legged robots in complex environments. 

In contrast to full-order models of legged locomotion, template models present reduced-order representations of legged robots that significantly reduce the computational burden and complexity associated with trajectory optimization. Various template models, including LIP \cite{kajita19991LIP}, SRB \cite{KF_mitcheetah3,Wensing_VBL_HJB,Abhishek_Hae-Won_TRO}, and centroidal dynamics \cite{orin2013centroidal}, have been successfully integrated with the MPC framework for the real-time planning of bipedal and quadrupedal robots \cite{Leonessa_Pratt_MPC,Ott_MPC,Pratt_LIP, dai2014whole,Abhishek_Hae-Won_TRO,Kim_Wensing_Convex_MPC_01, KF_mitcheetah3,Wensing_VBL_HJB,grandia2019frequency,hamed2020quadrupedal,Randy_Hamed_ASME,pandala2022robust}. The main challenge with using template models is bridging the gap between reduced- and full-order models of locomotion arising from abstraction (e.g., ignoring the legs' dynamics in template models). In particular, one needs to translate the optimal reduced-order trajectories to the full-order joint positions and torques. Different hierarchical control algorithms have been proposed in the literature to close this gap, in which a whole-body motion controller is utilized at the low level to map the optimal trajectories, generated by the higher-level MPC, to the full-order dynamics. For instance, \cite{Abhishek_Hae-Won_TRO,Wensing_VBL_HJB} have used Jacobian mapping, \cite{hamed2020quadrupedal,Jeeseop_Hamed_ASME} have used HZD-based controllers, \cite{pandala2022robust} has used robust MPC integrated with reinforcement learning, \cite{fawcett2022toward} has used data-driven template models, and \cite{farshidian2017efficient,bellicoso2016perception} have used joint space whole-body controllers. 

Despite the success of the above methods on individual robots,  it is unknown what reduced-order models can represent multi-agent-legged robots’ dynamic and cooperative transportation effectively.
In addition, it is unclear if the existing MPC techniques can address the real-time trajectory planning for the reduced-order models of cooperative locomotion with increased dimensionality. Moreover, it is unclear how the centralized MPC algorithms for such complex models can be decomposed into lower-dimensional distributed MPC algorithms considering the strong interaction terms. Our previous work in \cite{Jeeseop_Hamed_ASME} employed an interconnected network of LIP models with event-based MPC \cite{hamed2020quadrupedal} as a trajectory planner for cooperative locomotion. The simple nature of the LIP model and event-based MPC reduced the computational burden by running the MPC only at the beginning of the continuous-time domains rather than every time sample. However, using the LIP model prohibits us from capturing the interaction torques due to the assumption of a concentrated point mass at the center of mass (COM). This model also restricts the generation of dynamic cooperative gaits because the center of pressure (COP) must always remain within the support polygon, limiting the system's full potential. Moreover, the proposed event-based MPC was formulated only in a centralized manner and validated on numerical simulations and without experimental validations. In the current work, we aim to develop a new framework to allow more dynamic cooperative gaits while solving MPC problems faster in both centralized and distributed manners and experimentally validating the approach on two dynamic quadrupedal robots. 


\vspace{-0.5em}
\subsection{Objectives and Contributions}
\label{Sec:Objectives and Contributions} 

The \textit{objectives} and \textit{key contributions} of this paper are  as follows: 


\begin{enumerate}
    \item The paper presents an innovative network of two holonomically constrained SRB dynamics as an effective reduced-order model to capture the interaction wrenches between agents while dynamically stabilizing the motion during cooperative locomotion. It is numerically shown that the MPC algorithms utilizing a nominal SRB model cannot stabilize cooperative locomotion. 
    
    \item A layered control approach is proposed to robustly stabilize cooperative locomotion of holonomically constrained quadrupedal robots. At the high level of the control hierarchy, two different MPC algorithms, based on QP, are proposed: centralized MPC and distributed MPC (see Figs. \ref{fig:overview_centralized} and \ref{fig:overview_distributed}). The centralized MPC algorithm solves for the optimal state trajectory, GRFs, and interaction wrenches for the interconnected SRB dynamics. The distributed MPC algorithm assumes two local QPs that share their optimal solutions with a one-step communication delay. The distributed MPCs solve for the local states, local GRFs, and estimated local interaction wrenches according to an agreement protocol in the cost function.  

    \item At the low level of the proposed control architecture, distributed and full-order nonlinear controllers are presented for the whole-body motion control of agents. The distributed nonlinear controllers are developed based on QP and virtual constraints to impose the full-order dynamics to track the prescribed and optimal reduced-order trajectories and GRFs, generated by the high-level MPC (centralized or distributed).  
    
    \item Extensive numerical simulations are presented to evaluate the performance of the cooperative locomotion of two holonomically constrained A1 robots with different payloads on different rough terrains and in the presence of external force disturbances. A comparative analysis of the closed-loop systems with centralized and distributed MPC algorithms with more than 1000 randomly generated rough terrain profiles and external forces is presented. It is shown that the proposed distributed MPC algorithm has a performance similar to that of the centralized one, while the solve time is reduced by $70\%$. In addition, it is shown that the proposed centralized and distributed MPCs can drastically improve the robust stability of cooperative locomotion subject to a wide range of uncertainties, while the nominal MPCs cannot stabilize it. 
    
    \item The effectiveness of the proposed layered control algorithms (centralized and distributed) is verified with an extensive set of experiments for the blind and cooperative locomotion of two holonomically constrained A1 quadrupedal robots, each with 18 DOFs. The experiments include cooperative locomotion with different and unknown payloads on different terrains (covered with blocks, gravel, mulch, and slippery surfaces) and in the presence of external pushes and tethered pulling. Detailed robustness analysis is presented to experimentally evaluate the performance of the closed-loop system against the violations of assumptions made for the synthesis of the controller. 
\end{enumerate}

\vspace{-1.1em}
\subsection{Organization}
\label{Sec:Outline}

The paper is organized as follows. Section \ref{Sec:Reduced-Order Model of Cooperative Locomotion} develops interconnected SRB models as a reduced-order model of cooperative locomotion. Section \ref{Sec:Predictive Control based Planner} formulates centralized and distributed MPC-based trajectory planning algorithms with the proposed reduced-order model. Section \ref{Sec:Distributed Nonlinear Controllers for Full-Order Models} presents distributed nonlinear controllers for the whole-body motion control. Section \ref{Sec:Numerical and Experimental Validations} provides a detailed and extensive set of numerical and experimental validations of the proposed layer control algorithm. In Section \ref{Sec:Discussion and Comparison}, we discuss the results and compare the performance of the centralized and distributed MPC algorithms. Section \ref{Sec:Conclusion and Future Work} finally presents some concluding remarks and future research directions. 


\section{Reduced-Order Model of Cooperative Legged Locomotion}
\label{Sec:Reduced-Order Model of Cooperative Locomotion}

This section aims to address the reduced-order models that describe the cooperative locomotion of two holonomically constrained quadrupedal robots. The section assumes a rigid bar connected via ball joints to two points on the robots for carrying objects (see Fig. \ref{fig:2agents_snap}). These two points will be referred to as the \textit{interaction points}. This assumption simplifies the analysis and results in a holonomic constraint, stating that the Euclidean distance between the interaction points is constant. However, the analysis of this section can be extended to more sophisticated connections, such as restricting the pitch or roll angles of the bar/load. In Section \ref{subsec:rotationregulation}, we will experimentally show the robustness of the developed algorithms subject to these additional constraints. 

\begin{figure}[t!]
\centering
\includegraphics[width=\linewidth]{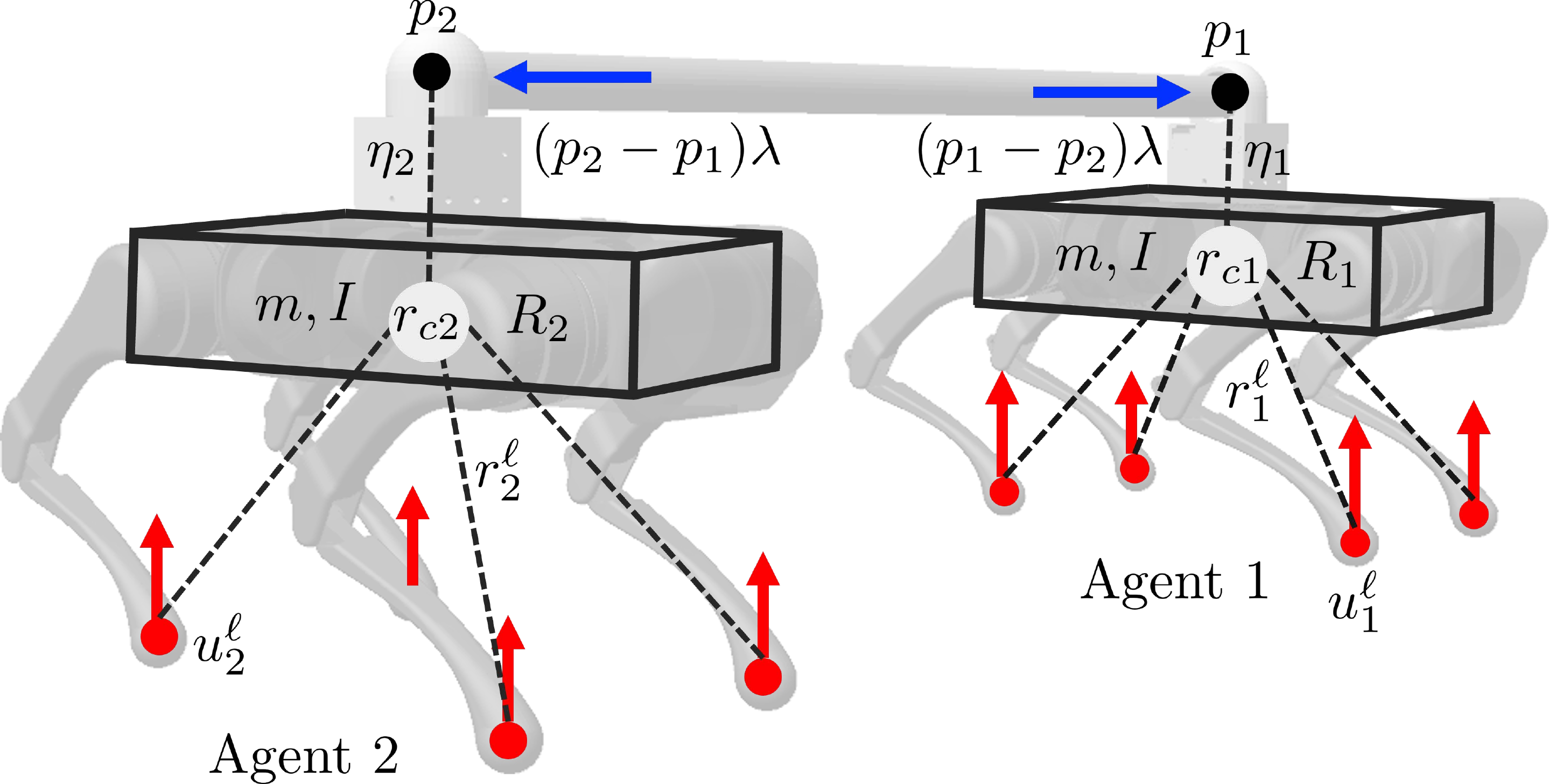}
\vspace{-1.5em}
\caption{Illustration of the interconnected SRB models for the cooperative locomotion of two quadrupedal robots.}
\label{fig:interconnected_srb}
\vspace{-1.5em}
\end{figure}

In our notation, the subscript $i\in\mathcal{I}:=\{1,2\}$ represents the $i$th robot. We assume that $\{B_{i}\}$ is the local frame rigidly attached to the body of the agent $i$ with its origin on the COM. The orientation of the frame $\{B_{i}\}$ with respect to the inertial world frame $\{O\}$ is denoted by $R_{i}\in\textrm{SO}(3)$, where $\textrm{SO}(3):=\{R\in\Real^{3\times3}\,|\,R^\top R=\iden, \textrm{det}(R)=1\}$ is the special orthogonal group of order $3$, and $\iden$ represents the identity matrix. The Cartesian coordinates of the COM of agent $i$ with respect to $\{O\}$ are also represented by $r_{ci}:=\col(x_{ci},y_{ci},z_{ci})\in\Real^{3}$, where ``$\col$'' denotes the column operator. Moreover, $\omega_{i}^{B_{i}}\in\Real^{3}$ represents the angular velocity of agent $i$ expressed in the body frame $\{B_{i}\}$. We assume that $p_{i}\in\Real^{3}$ for $i\in\mathcal{I}$ represents Cartesian coordinates of the interaction points with respect to the inertial frame $\{O\}$, that is, 
\begin{equation}\label{interaction_points}
    p_{i} = r_{ci} + R_{i}\,d_{i}^{B_{i}},
\end{equation}
where $d_{i}^{B_{i}}\in\Real^{3}$ is a constant vector denoting the coordinates of the interaction points in the body frame $\{B_{i}\}$. For future purposes, we define $\eta_{i}:=R_{i}\,d_{i}^{B_{i}}$ (see Fig. \ref{fig:interconnected_srb}). We remark that the holonomic constraint between two agents can be described as a constraint on the Euclidean distance between the interaction points as follows: 
\begin{equation}\label{holonomic_constraint}
    \psi\left(r_{c1},r_{c2},R_{1},R_{2}\right):=\frac{1}{2}\left\|p_{1} - p_{2}\right\|^{2} = \psi_{0},
\end{equation}
in which $\|\cdot\|$ denotes the $2$-norm, and $\psi_{0}$ is a constant number, determined based on the length of the bar.

According to the principle of virtual work, one can consider $(p_{1}-p_{2})\,\lambda\in\Real^{3}$ as the interaction force applied to agent $1$ for some Lagrange multiplier $\lambda\in\Real$ to be determined later (see again Fig. \ref{fig:interconnected_srb}). Consequently, the net external wrench applied to agent $i\in\mathcal{I}$ can be expressed as follows:
\begin{equation}\label{net_wrench}
    \begin{bmatrix}
    f_{i}^{\net}\\
    \tau_{i}^{\net}
    \end{bmatrix}=\sum_{\ell\in\mathcal{C}_{i}}\begin{bmatrix}
    \iden\\
    \widehat{r}_{i}^{\ell}
    \end{bmatrix}u_{i}^{\ell} + \begin{bmatrix}
    \iden\\
    \widehat{\eta_{i}}
    \end{bmatrix}\left(p_{i} - p_{j}\right)\lambda,
\end{equation}
where $j\neq i\in\mathcal{I}$ denotes the index of the other agent and  the hat map $\widehat{(\cdot)}: \Real^3 \rightarrow \mathfrak{so}(3)$ represents the skew-symmetric operator with the property $\widehat{x}\,y=x\times y$ for all $x,y\in\Real^{3}$. In \eqref{net_wrench}, the superscript $\ell\in\mathcal{C}_{i}$ denotes the index of the contacting legs with the ground,  $\mathcal{C}_{i}$ represents the set of contacting legs for the agent $i$, and $u^{\ell}_{i}\in\Real^{3}$ denotes the GRF at the contacting leg $\ell$ for the agent $i$. In addition, $r_{i}^{\ell}\in\Real^{3}$ represents the position of each contacting leg with respect to the COM of agent $i$, that is, 
$r_{i}^{\ell} = r_{\textrm{foot},i}^{\ell} - r_{ci}$, where $r_{\textrm{foot},i}^{\ell}$ is the position of the contacting foot $\ell$ of the agent $i$ with respect to the world frame $\{O\}$. 

By taking the \textit{local state variables} for the agent $i\in\mathcal{I}$ as
\begin{equation}
    \bm{x}_{i}:=\col\left(r_{ci},\dot{r}_{ci},\vect(R_{i}),\omega_{i}^{B_{i}}\right)\in\Real^{18},
\end{equation}
the \textit{global state variables} can be defined as
\begin{equation}
    \bm{x}:=\col\left(\bm{x}_{1},\bm{x}_{2}\right),
\end{equation}
where ``$\vect$'' represents the vectorization operator. Similarly, the \textit{global control inputs} can be defined as $\bm{u}:=\col(\bm{u}_{1},\bm{u}_{2})$, where $\bm{u}_{i}$ denotes the \textit{local control inputs} (i.e., GRFs) for the agent $i$, that is,
\begin{equation}
    \bm{u}_{i}:=\col\left\{u_{i}^{\ell}\,|\,\ell\in\mathcal{C}_{i}\right\}.
\end{equation}
By differentiating the holonomic constraint \eqref{holonomic_constraint}, one can get
\begin{equation}\label{1st_order_derivative_holonomic_constraint}
    \dot{\psi}(\bm{x})=\left(p_{1} - p_{2}\right)^\top \left(\dot{p}_{1} - \dot{p}_{2}\right)=0,
\end{equation}
and hence, the \textit{state manifold} for the interconnected SRB dynamics can be expressed as 
\begin{alignat}{4}
    \bm{\mathcal{X}}\!:=\!\left\{\bm{x}\in\Real^{36}|\,R_{i}\in\textrm{SO}(3),\,i\in\mathcal{I},\,\psi(\bm{x})=\psi_{0},\dot{\psi}(\bm{x})=0\right\}\!.\nonumber
\end{alignat}
Finally, the interconnected SRB dynamics can be expressed as
\begin{equation}\label{interconnected_SRB_NLdyn}
    \dot{\bm{x}}=\bm{f}(\bm{x},\bm{u},\lambda):=\begin{bmatrix}
    \dot{r}_{c1}\\
    \frac{f_{1}^{\net}}{m} - g\\
    \vect(R_{1}\,\widehat{\omega_{1}^{B_{1}}})\\
    I^{-1}\left(R_{1}^\top\tau^{\net}_{1} - \widehat{\omega_{1}^{B_{1}}}\,I\,\omega_{1}^{B_{1}}\right)\\
    \dot{r}_{c2}\\
    \frac{f_{2}^{\net}}{m} - g\\
    \vect(R_{2}\,\widehat{\omega_{2}^{B_{2}}})\\
    I^{-1}\left(R_{2}^\top\tau^{\net}_{2} - \widehat{\omega_{2}^{B_{2}}}\,I\,\omega_{2}^{B_{2}}\right)
    \end{bmatrix},
\end{equation}
where $m$ and $I\in\Real^{3\times3}$ denote the total mass and the fixed moment of inertia in the body frame for each agent, respectively, and $g$ represents the constant gravitational vector. 
We remark that the kinematics relations in \eqref{interconnected_SRB_NLdyn} are expressed as $\dot{R}_{i}=R_{i}\,\widehat{\omega_{i}^{B_{i}}}$ for $i\in\mathcal{I}$. The rotational dynamics can be further expressed as Euler's equation $I\,\dot{\omega}_{i}^{B_{i}}+\widehat{\omega_{i}^{B_{i}}}\,I\,\omega_{i}^{B_{i}}=R_{i}^\top\tau_{i}^{\net}$. We also note that in \eqref{interconnected_SRB_NLdyn}, $\bm{f}:\bm{\mathcal{X}}\times\bm{\mathcal{U}}\times\Real\rightarrow \textrm{T}\bm{\mathcal{X}}$ is smooth with 
\begin{equation}\bm{\mathcal{U}}:=\underbrace{\mathcal{FC}\times\cdots\times\mathcal{FC}}_{m_{u}- \textrm{times}}\subset\Real^{3m_{u}}
\end{equation}
being the \textit{admissible set of control inputs}, where $m_{u}$ denotes the total number of contacting legs with the ground (e.g., $m_{u}=4$ for cooperative trot), $\mathcal{FC}:=\{\col(f_{x},f_{y},f_{z})\,|\,f_{z}>0,\,|f_{x}|\leq\frac{\mu}{\sqrt{2}} f_{z},\,|f_{y}|\leq\frac{\mu}{\sqrt{2}} f_{z}\}$ represents the linearized friction cone for some friction coefficient $\mu$, and $\textrm{T}\bm{\mathcal{X}}$ is the tangent bundle of the state manifold $\bm{\mathcal{X}}$. 

In order to make the manifold $\bm{\mathcal{X}}$ invariant under the flow of \eqref{interconnected_SRB_NLdyn}, one would need to choose the Lagrange multiplier $\lambda$ to satisfy the holonomic constraint. In particular, differentiating \eqref{1st_order_derivative_holonomic_constraint} according to \eqref{interaction_points} and $\dot{R}_{i}=R_{i}\,\widehat{\omega_{i}^{B_{i}}}$ results in 
\begin{alignat}{4}
    &\ddot{\psi}(\bm{x},\bm{u},\lambda)&&=\left(p_{1} - p_{2}\right)^\top \left(\ddot{p}_{1} - \ddot{p}_{2}\right)+\left\|\dot{p}_{1}-\dot{p}_{2}\right\|^{2}\nonumber\\
    & &&=\left(p_{1}-p_{2}\right)^\top\Big\{\ddot{r}_{c1}-\ddot{r}_{c2}\nonumber\\
    & &&+R_{1}\left(\widehat{\omega_{1}^{B_{1}}}\right)^{2}d_{1}^{B_{1}}-R_{2}\left(\widehat{\omega_{2}^{B_{2}}}\right)^{2}d_{2}^{B_{2}}\nonumber\\
    & &&+R_{1}\,\widehat{\dot{\omega}_{1}^{B_{1}}}d_{1}^{B_{1}} - R_{2}\,\widehat{\dot{\omega}_{2}^{B_{2}}}d_{2}^{B_{2}}\Big\}\nonumber\\
    & &&+\left\|\dot{p}_{1}-\dot{p}_{2}\right\|^{2}=0.\label{2nd_order_derivative_holonomic_constraint}
\end{alignat}
This latter equation, together with the equations of motion \eqref{interconnected_SRB_NLdyn} and \eqref{net_wrench}, results in $\lambda$ being a function of $(\bm{x},\bm{u})$. However, replacing this nonlinear expression for $\lambda$ in \eqref{interconnected_SRB_NLdyn} can make the original dynamics \eqref{interconnected_SRB_NLdyn} more nonlinear and complex. Furthermore, this can numerically complicate the Jacobian linearization of $\dot{\bm{x}}=\bm{f}(\bm{x},\bm{u},\lambda(\bm{x},\bm{u}))$ when formulating the trajectory planning problem as a convex MPC in Section \ref{Sec:Predictive Control based Planner}. Alternatively, we pursue a computationally effective approach by considering $\dot{\bm{x}}=\bm{f}(\bm{x},\bm{u},\lambda)$ subject to the equality constraint $\ddot{\psi}(\bm{x},\bm{u},\lambda)=0$ within the optimal control problem formulation. More specifically, the decision variables for the MPC include the trajectories of $(\bm{x},\bm{u},\lambda)$ over the control horizon, and the MPC will satisfy the equality constraint. The other advantage of this technique is that the interconnected SRB dynamics can be integrated with the variational-based approach of \cite{Abhishek_Hae-Won_TRO,Wensing_VBL_HJB} to linearize and then discretize the dynamics such that the rotation matrices $R_{i}, i\in\mathcal{I}$ evolve on $\textrm{SO}(3)$. 

To clarify this latter point, following \cite{Abhishek_Hae-Won_TRO}, we introduce a new set of local state variables for the agent $i\in\mathcal{I}$ with the abuse of notation as 
\begin{equation}
\bm{x}_{i}:=\col\left(r_{ci},\dot{r}_{ci},\xi_{i},\omega^{B_{i}}_{i}\right)\in\Real^{12}.
\end{equation}
Here, $\xi_{i}\in\Real^{3}$ is a vector used to approximate the rotation matrix $R_{i}$ around an operating point $R_{i,\op}$ as follows:
\begin{equation}\label{R_approx}
    R_{i} =R_{i,\op}\,\exp(\widehat{\xi_{i}}) \approx R_{i,\op}\left(\iden+\widehat{\xi_{i}}\right).
\end{equation}
The approach of \cite{Abhishek_Hae-Won_TRO} has linearized the SRB dynamics subject to GRFs without interaction forces. Hence, one must extend the technique to write down the Taylor series expansion for the additional wrench terms in \eqref{net_wrench} arising from the interaction. This results in a discrete and linear time-varying (LTV) system to predict the future states as follows:
\begin{equation}\label{prediction_dyn}
    \bm{x}_{k+t+1|t} = \bm{A}_{\op}\,\bm{x}_{k+t|t} + \bm{B}_{\op}\,\bm{u}_{k+t|t} + \bm{C}_{\op}\, \lambda_{k+t|t} + \bm{d}_{\op},
\end{equation}
for all $k=0,1,\cdots,N-1$ and with the initial condition $\bm{x}_{t|t}=\bm{x}_{t}$. Here, $\bm{x}\in\Real^{24}$ denotes the global state variables, $N$ represents the control horizon, and $(\bm{x}_{k+t|t},\bm{u}_{k+t|t},\lambda_{k+t|t})$ denotes the tuple of the predicted global states, global inputs (i.e., GRFs), and Lagrange multiplier at time $k+t$ computed at time $t$. Furthermore, $\bm{A}_{\op}\in\Real^{24\times24}$, $\bm{B}_{\op}\in\Real^{24\times 3m_{u}}$, $\bm{C}_{\op}\in\Real^{24}$, and $\bm{d}_{\op}\in\Real^{24}$ are the Jacobian matrices and offset term evaluated around the current operating point $(\bm{x}_{t},\bm{u}_{t-1},\lambda_{t-1})$. 

The approximation in \eqref{R_approx} only ensures that the rotation matrices evolve on $\textrm{SO}(3)$. To guarantee that the state predictions in \eqref{prediction_dyn} belong to the tangent space of the state manifold at the operating point (i.e., $\textrm{T}_{\op}\bm{\mathcal{X}}$), we first define the following equality constraint  
\begin{equation}\label{NL_eq}
    \bm{\Psi}(\bm{x},\bm{u},\lambda):=\begin{bmatrix}
    \psi(\bm{x})-\psi_{0}\\
    \dot{\psi}(\bm{x})\\
    \ddot{\psi}(\bm{x},\bm{u},\lambda)
    \end{bmatrix}=0.
\end{equation}
Then, analogous to the technique used for the linearization of the interconnected dynamics, the equality constraint \eqref{NL_eq} can be approximated around the operating point as follows:
\begin{equation}\label{equality_constrinats}
    \bm{E}_{\op}\,\bm{x}_{k+t|t} + \bm{F}_{\op}\,\bm{u}_{k+t|t} + \bm{G}_{\op}\,\lambda_{k+t|t} + \bm{h}_{\op} = 0
\end{equation}
to ensure that $\bm{\Psi}(\bm{x}_{k+t|t},\bm{u}_{k+t|t},\lambda_{k+t|t})\equiv0$. Here, $\bm{E}_{\op}\in\Real^{3\times24}$, $\bm{F}_{\op}\in\Real^{3\times 3m_{u}}$, $\bm{G}_{\op}\in\Real^{3}$, and $\bm{h}_{\op}\in\Real^{3}$ are proper matrices and vectors that can be either computed via the approach of \cite{Abhishek_Hae-Won_TRO} or symbolic calculus.  

\begin{remark} 
As the nature of the holonomic constraints between the agents becomes more complex, the procedure for obtaining the corresponding prediction model and equality constraints becomes computationally expensive. However, our experimental results in Section \ref{subsec:rotationregulation} will indicate that the proposed layered control approach, developed based on the assumption of holonomic constraints in \eqref{holonomic_constraint}, can robustly stabilize cooperative locomotion subject to uncertainties in the constraints (e.g., limiting the pitch angles of the ball joints). In addition, Section \ref{Sec:Numerical Validation} will show that ignoring the holonomic constraints \eqref{holonomic_constraint} for the reduced-order model and trajectory planner can \textit{destabilize} cooperative locomotion.  
\end{remark}


\section{MPC-Based Trajectory Planning}
\label{Sec:Predictive Control based Planner}

This section aims to formulate the real-time trajectory planning problem for cooperative locomotion as centralized and distributed MPC algorithms. 


\vspace{-0.5em}
\subsection{Centralized MPC}
\label{Sec:Centralized Model Predictive Control}

We will consider a locomotion pattern for the agents, described by the directed cycle  $\mathcal{G}(\mathcal{V},\mathcal{E})$, where $\mathcal{V}$ and $\mathcal{E}\subset\mathcal{V}\times\mathcal{V}$ represent the sets of vertices and edges, respectively. The vertices denote the continuous-time domains of locomotion, and the edges represent the discrete-time transitions between the continuous-time domains. 

\begin{assumption}\label{Assumption:Synchronization}
At every time sample $t$, the higher-level MPC is aware of the current stance legs, assuming that the stance leg configuration does not change throughout the prediction horizon.
\end{assumption}

\begin{remark}\label{Remark:synchrony}
Assumption \ref{Assumption:Synchronization} is \textit{not} restrictive and simplifies the optimal control problem of \eqref{prediction_dyn} subject to \eqref{equality_constrinats} over the control horizon. Otherwise, one would need to consider the optimal control problem for a piecewise affine (PWA) system \cite[Chap. 16]{MPC_Book} subject to different switching times. 
\end{remark}

We are now in a position to present the following real-time centralized MPC algorithm for the cooperative locomotion
\begin{alignat}{4}
&\min_{(\bm{x}(\cdot),\bm{u}(\cdot),\lambda(\cdot))} &&p\left(\bm{x}_{t+N|t}\right) + \sum_{k=0}^{N-1} \mathcal{L}\left(\bm{x}_{k+t|t},\bm{u}_{k+t|t},\lambda_{k+t|t}\right)\nonumber\\
&\quad\quad\textrm{s.t.}&&\textrm{Prediction model \eqref{prediction_dyn}}\nonumber\\
& &&\textrm{Equality constraints \eqref{equality_constrinats}}\nonumber\\
& && \bm{u}_{k+t|t}\in\bm{\mathcal{U}}, \quad k=0,1,\cdots,N-1,\label{centralized_MPC}
\end{alignat}
where the equality constraints for the MPC arise from a) the prediction model \eqref{prediction_dyn} to address the interconnected SRB dynamics with the initial condition of $\bm{x}_{t|t}=\bm{x}_{t}$, and b) the holonomic constraints \eqref{equality_constrinats} (see Fig. \ref{fig:overview_centralized}). Here, the centralized MPC solves for the optimal trajectories of the global states, global inputs, and the Lagrange multiplier encoded in $(\bm{x}(\cdot),\bm{u}(\cdot),\lambda(\cdot))$ to retain the sparsity structure of \cite{Boyd_FastMPC}, where $\bm{x}(\cdot):=\col\{\bm{x}_{k+t|t}\,|\,k=1,\cdots,N\}$, $\bm{u}(\cdot):=\col\{\bm{u}_{k+t|t}\,|\,k=0,1,\cdots,N-1\}$, and $\lambda(\cdot):=\col\{\lambda_{k+t|t}\,|\,k=0,1,\cdots,N-1\}$. The terminal and stage cost functions in \eqref{centralized_MPC} are then taken as $p(\bm{x}_{t+N|t}):=\|\bm{x}_{t+N|t}-\bm{x}^{\des}_{t+N|t}\|_{\bm{P}}^{2}$ and $\mathcal{L}(\bm{x}_{k+t|t},\bm{u}_{k+t|t},\lambda_{k+t|t}):=\|\bm{x}_{k+t|t} - \bm{x}^{\des}_{k+t|t}\|_{\bm{Q}}^{2} + \|\bm{u}_{k+t|t} \|_{\bm{R}_{u}}^{2} + \|\lambda_{k+t|t}\|_{\bm{R}_{\lambda}}^{2}$ for some desired trajectory $\bm{x}^{\des}(\cdot)$ and some positive definite matrices $\bm{Q}$ and $\bm{R}_{u}$, and a positive scalar $\bm{R}_{\lambda}$. Finally, the inequality constraints of \eqref{centralized_MPC} represent the feasibility of the GRFs for two agents. 
\begin{remark}
The MPC in \eqref{centralized_MPC} addresses the trajectory planning problem over the current continuous-time domain. In particular, we do not include the following domain for prediction purposes. This is mainly due to the fact that the actual footholds for the following domain are \textit{not} known \textit{a priori}. More specifically, we employ Raibert’s heuristic \cite[Eq. (2.4), pp. 46]{raibert1986legged} to plan for the upcoming footholds of each agent.  Assuming pre-planned footholds, one can extend the MPC to include other domains. However, our experimental results in Section \ref{Sec:Numerical and Experimental Validations} suggest that planning over the current domain is sufficient for robustly stable cooperative locomotion. This is in agreement with most of the existing MPC approaches for single SRB dynamics. We also remark that the centralized MPC has $(25+3\,m_{u})\,N$ decision variables, where $m_{u}$ represents the total number of contacting legs with the ground.  
Finally, the MPC problem \eqref{centralized_MPC} solves for the optimal trajectories of the state variables $\bm{x}^{\star}(\cdot)$, control inputs $\bm{u}^{\star}(\cdot)$, and Lagrange multiplier $\lambda^{\star}(\cdot)$. However, the high-level MPC only applies the first element of the optimal state and control sequence, i.e., $(\bm{x}^{\star}_{t+1|t},\bm{u}^{\star}_{t|t})$, to the low-level nonlinear controller for tracking while discarding $\lambda_{t|t}^{\star}$ (see Fig. \ref{fig:overview_centralized}). 
\end{remark}


\vspace{-0.5em}
\subsection{Distributed MPC}
\label{Sec:Distributed Model Predictive Control}

This section aims to develop a network of distributed MPCs with a smaller number of decision variables that plan for the cooperative locomotion of two holonomically constrained quadrupedal robots. From \eqref{prediction_dyn}, the local dynamics of the agent $i\in\mathcal{I}$ can be expressed as follows:
\begin{alignat}{4}
& \bm{x}_{i,k+t+1|t} && =\bm{A}_{\op,ii}\,\bm{x}_{i,k+t|t} + \bm{B}_{\op,ii}\,\bm{u}_{i,k+t|t} \nonumber\\
& && + \bm{C}_{\op,i}\,\lambda_{k+t|t} + \bm{d}_{\op,i} +  \bm{\Delta}_{i}\left(\bm{x}_{j,k+t|t},\bm{u}_{j,k+t|t}\right),\label{local_dyn}
\end{alignat}
for $j\neq i\in\mathcal{I}$, where $\bm{A}_{\op,ii},\bm{A}_{\op,ij}\in\Real^{12\times12}$, $\bm{B}_{\op,ii},\bm{B}_{\op,ij}\in\Real^{12\times\frac{3}{2}m_{u}}$, $\bm{C}_{\op,i}\in\Real^{12}$, and $\bm{d}_{\op,i}\in\Real^{12}$ denote the corresponding partitioning of $(\bm{A}_{\op},\bm{B}_{\op},\bm{C}_{\op},\bm{d}_{\op})$. In addition, 
\begin{equation}\label{local_interaction}
    \bm{\Delta}_{i}\left(\bm{x}_{j,k+t|t},\bm{u}_{j,k+t|t}\right):=\bm{A}_{\op,ij}\,\bm{x}_{j,k+t|t} + \bm{B}_{\op,ij}\,\bm{u}_{j,k+t|t}
\end{equation}
represents the interaction term on the agent $i$. Similarly, the equality constraints \eqref{equality_constrinats} can be rewritten as follows:
\begin{alignat}{4}
& \bm{E}_{\op,i}\,\bm{x}_{i,k+t|t} && + \bm{F}_{\op,i}\,\bm{u}_{i,k+t|t} + \bm{G}_{\op} \lambda_{k+t|t} + \bm{h}_{\op}\nonumber\\
& && +\bm{\Omega}_{i}\left(\bm{x}_{j,k+t|t},\bm{u}_{j,k+t|t}\right)=0,\label{local_equality_constrinats}
\end{alignat}
in which $\bm{E}_{\op,i}\in\Real^{3\times12}$ and $\bm{F}_{\op,i}\in\Real^{3\times\frac{3}{2}m_{u}}$ are the corresponding columns of $(\bm{E}_{\op},\bm{F}_{\op})$, and 
\begin{equation*}
    \bm{\Omega}_{i}\left(\bm{x}_{j,k+t|t},\bm{u}_{j,k+t|t}\right):=\bm{E}_{\op,j}\,\bm{x}_{j,k+t|t} + \bm{G}_{\op,j}\,\bm{u}_{j,k+t|t}
\end{equation*}
for $j\neq i$. Motivated by the inherent limitation of the distributed QP problems, one would need to estimate the interaction terms $\bm{\Delta}_{i}$ and $\bm{\Omega}_{i}$, $i\in\mathcal{I}$ to solve for local QPs. For this purpose, we make the following assumption. 

\begin{assumption}[One-Step Communication Protocol]\label{Assumption:Communication_Protocol}
At every time sample $t$, each local MPC has access to the optimal solution of the other local MPC at time $t-1$. More specifically, the local MPCs share their previous optimal solutions over the network.
\end{assumption}

From Assumption \ref{Assumption:Communication_Protocol}, we can estimate the interaction terms $\bm{\Delta}_{i}$ and $\bm{\Omega}_{i}$ in \eqref{local_dyn} and \eqref{local_equality_constrinats} using the previous optimal solutions, that is,
\begin{alignat}{4}
    & \bm{\Delta}_{i}\left(\bm{x}_{j,k+t|t},\bm{u}_{j,k+t|t}\right) && \approx
    \bm{\Delta}_{i}\left(\bm{x}_{j,k+t|t-1}^{\star},\bm{u}_{j,k+t|t-1}^{\star}\right)\nonumber\\
    & \bm{\Omega}_{i}\left(\bm{x}_{j,k+t|t},\bm{u}_{j,k+t|t}\right) && \approx
    \bm{\Omega}_{i}\left(\bm{x}_{j,k+t|t-1}^{\star},\bm{u}_{j,k+t|t-1}^{\star}\right),\label{estimation_of_intercation}
\end{alignat}
in which $\bm{x}_{j,k+t|t-1}^{\star}$ and $\bm{u}_{j,k+t|t-1}^{\star}$ denote the optimal solution from the local QP $j$ for time $k+t$ computed at time $t-1$ for $k=0,1,\cdots,N-1$. We remark that as the QP $j$ does not plan for $\bm{u}_{N-1+t|t-1}$, we let $\bm{u}_{N-1+t|t-1}^{\star}=0$. The assumption in \eqref{estimation_of_intercation} estimates the interaction terms in the local dynamics and equality constraints based on the optimal values from the local QP $j$ at the previous time sample. With this assumption, the local MPC $i$ needs to optimally solve for its own local state trajectory $\bm{x}_{i}(\cdot)$, local control trajectory $\bm{u}_{i}(\cdot)$, and the Lagrange multiplier trajectory $\lambda(\cdot)$. However, as the Lagrange multiplier $\lambda$ is common between the decision variables of two local MPCs, they need to reach a consensus over time for the optimal $\lambda$ value. 

To address the consensus problem, we develop an agreement protocol as follows. The cost function of the centralized MPC \eqref{centralized_MPC} can be written as the sum of individual terms, i.e., 
\begin{equation}
    \mathcal{J}_{1}\left(\bm{x}_{1}(\cdot),\bm{u}_{1}(\cdot)\right) + \mathcal{J}_{2}\left(\bm{x}_{2}(\cdot),\bm{u}_{2}(\cdot)\right) +
    \mathcal{J}_{\lambda}\left(\lambda(\cdot)\right).
\end{equation}
We assume that each local QP estimates its own trajectory of the Lagrange multiplier, denoted by $\lambda_{i}(\cdot)$. We then propose the following real-time distributed MPC for agent $i\in\mathcal{I}$
\begin{alignat}{4}
& \min_{(\bm{x}_{i}(\cdot),\bm{u}_{i}(\cdot),\lambda_{i}(\cdot))} && \mathcal{J}_{i}\left(\bm{x}_{i}(\cdot),\bm{u}_{i}(\cdot)\right) + \mathcal{J}_{\lambda}\left(\lambda_{i}(\cdot)\right) \nonumber\\
& && \!\!\!\!\!\!\!\!\!\!\!\!\!\!\!\!\!\!\!+ w \sum_{k=0}^{N-1} \| \lambda_{i,k+t|t} - a_{ii}\,\lambda_{i,k+t|t-1}^{\star} - a_{ij} \,\lambda_{j,k+t|t-1}^{\star} \|^{2} \nonumber\\
& && \!\!\!\!\!\!\!\!\!\!\!\!\!\!\!\!\!\!\!+ \beta_{j,t-1}^{\star \top}\,\mathcal{K}_{j,i} \begin{bmatrix}
\bm{x}_{i}(\cdot)\\
\bm{u}_{i}(\cdot)
\end{bmatrix} + \beta_{j,t-1}^{\star \top}\,\mathcal{K}_{j,\lambda}\, \lambda_{i}(\cdot)\nonumber\\
&  \qquad\qquad \textrm{s.t.} &&
\textrm{Local prediction model \eqref{local_dyn} with \eqref{estimation_of_intercation}}\nonumber\\
& && \textrm{Local equality constraints \eqref{local_equality_constrinats} with \eqref{estimation_of_intercation}}\nonumber\\
& && \bm{u}_{i,k+t|t}\in\bm{\mathcal{U}_{i}}, \quad k=0,1,\cdots,N-1,\label{distributed_MPC}
\end{alignat}
where $w$ is a positive weighting factor added to introduce a new term in the local cost function to address the agreement protocol. In particular, the agreement term penalizes the difference between the local predicted values of $\lambda_{i,k+t|t}$ and the average of the previously computed optimal values $\lambda_{i,k+t|t-1}^{\star}$ and $\lambda_{j,k+t|t-1}^{\star}$ by the local MPCs $i$ and $j$ at time $t-1$. Here, $a_{ii}$ and $a_{ij}$ are the weighting factors for averaging with the property $a_{ii},a_{ij}\in[0,1]$ and $a_{ii} + a_{ij} = 1$, where $i\neq j\in\mathcal{I}$. The last two terms in the cost functions will be described shortly. The distributed MPC \eqref{distributed_MPC} has two sets of equality constraints arising from a) the local dynamics \eqref{local_dyn}, and b) the holonomic constraint \eqref{local_equality_constrinats} with the assumption \eqref{estimation_of_intercation}. 

The proposed local MPC for the agent $i$ does not consider the local dynamics of the other agent (i.e., agent $j$). Instead, motivated by our previous work \cite{Vinay_LCSS_distributedQPs}, it uses the Karush–Kuhn–Tucker (KKT) Lagrange multipliers that correspond to the equality constraint arising from the local dynamics of the agent $j$ in the QP $j$ at time $t-1$. This set of \textit{KKT Lagrange multipliers} is denoted by $\beta_{j,t-1}^{\star}$. In addition, $\mathcal{K}_{j,i}$ and $\mathcal{K}_{j,\lambda}$ represent the \textit{sensitivity} (i.e., gradient) of the local dynamics $j$ with respect to the local variables $(\bm{x}_{i}(\cdot),\bm{u}_{i}(\cdot))$ and $\lambda(\cdot)$, respectively. In particular, $\mathcal{K}_{j,i}$ can be computed in a straightforward manner by taking the gradient of the local interaction terms $\bm{\Delta}_{j}$ with respect to $(\bm{x}_{i,k+t|t},\bm{u}_{i,k+t|t})$ over the entire control horizon and stacking the results together, that is, 
\begin{alignat}{4}
&\mathcal{K}_{j,i}&&:=\frac{\partial\,\col\{\bm{\Delta}_{j}(\bm{x}_{i,k+t|t},\bm{u}_{i,k+t|t})\,|\,k=0,1,\cdots,N-1\}}
{\partial(\bm{x}_{i}(\cdot),\bm{u}_{i}(\cdot))}\nonumber.
\end{alignat}
An analogous approach can be used to compute the sensitivity matrix $\mathcal{K}_{j,\lambda}$. We then add the last two linear terms to the cost function of the local MPC \eqref{distributed_MPC}. Our previous work \cite[Theorem 1]{Vinay_LCSS_distributedQPs} has shown that the inclusion of the KKT Lagrange multipliers $\beta^{\star}_{j,t-1}$ together with the sensitivity matrices $(\mathcal{K}_{j,i},\mathcal{K}_{j,\lambda})$ in the cost function can result in a set of local KKT conditions that have a similar structure to that of the KKT equations for the centralized problem. Finally, $\bm{\mathcal{U}}_{i}$ in \eqref{distributed_MPC} represents the local feasibility set for the GRFs (i.e., inputs). 

\begin{remark}
We remark that local MPCs in the proposed structure \eqref{distributed_MPC} share their optimal local state trajectory $\bm{x}_{i}^{\star}(\cdot)$, local control trajectory $\bm{u}_{i}^{\star}(\cdot)$, local estimates of the Lagrange multiplier trajectory $\lambda_{i}^{\star}(\cdot)$, and the KKT Lagrange multipliers corresponding to the local dynamics in the QP structure $\beta_{i}^{\star}$ with the other agent and according to the one-step communication delay protocol (see Fig. \ref{fig:overview_distributed}). Finally, the number of decision variables for each local MPC becomes $(13+\frac{3}{2}\,m_{u})\,N$. Section \ref{subsec:lambdaevolution} will numerically study and show the consensus problem of the estimated Lagrange multipliers. 
\end{remark}


\section{Distributed Nonlinear Controllers for Full-Order Models}
\label{Sec:Distributed Nonlinear Controllers for Full-Order Models}

The objective of this section is to present the low-level and distributed nonlinear controllers for the whole-body motion control of each agent. The full-order and floating-based model of the agent $i$ can be described by the Euler-Lagrange equations and principle of virtual work as follows:
\begin{alignat}{4}
    & D({q}_{i})\,\ddot{{q}}_{i} + H({q}_{i},\dot{{q}}_{i}) && =\Upsilon\,{\tau}_{i} + \sum_{\ell\in\mathcal{C}_{i}} J_{\ell}^\top({q}_{i})\,f_{i}^{\ell}\nonumber\\  
    & && +J_{\inter}^{\top}({q}_{i})\,(p_{i}-p_{j})\,\lambda,\label{full_order_model}
\end{alignat}
where ${q}_{i}\in\mathcal{Q}\subset\Real^{n_{q}}$ represents the generalized coordinates of the robot $i$, $\mathcal{Q}$ and $n_{q}$ denote the configuration space and number of DOFs, respectively, ${\tau}_{i}\in\mathcal{T}\subset\Real^{n_{\tau}}$ represents the joint-level torques at the actuated joints, $\mathcal{T}$ is a closed and convex set of admissible torques, and $\mathcal{C}_{i}$ represents the set of contacting legs with the environment. In addition, $f_{i}^{\ell}$ denotes the GRF at the contacting leg $\ell\in\mathcal{C}_{i}$ of the full-order model for the agent $i$. We remark that the GRF at the contacting leg $\ell\in\mathcal{C}_{i}$ of the reduced-order model for the agent $i$ was denoted by $u_{i}^{\ell}$ in Section \ref{Sec:Reduced-Order Model of Cooperative Locomotion}. This is due to the possible gap between the reduced- and full-order models arsing from abstraction (i.e., ignoring legs' dynamics).
Moreover, $D({q}_{i})\in\Real^{n_{q}\times n_{q}}$ denotes the positive definite mass-inertia matrix, $H({q}_{i},\dot{{q}}_{i})\in\Real^{n_{q}}$ represents the Coriolis, centrifugal, and gravitational terms, $\Upsilon\in\Real^{n_{q}\times n_{\tau}}$ is the input distribution matrix, $J_{\ell}({q}_{i})$ denotes the contact Jacobin matrix at the leg $\ell$, $J_{\inter}({q}_{i})$ represents the Jacobian of the interaction point $p_{i}$, and $(p_{i}-p_{j})\,\lambda$ denotes the interaction force between the two agents as described in the reduced-order model of Section \ref{Sec:Reduced-Order Model of Cooperative Locomotion}. The local and full-order state variables for the agent $i$ is defined as ${z}_{i}:=\col({q}_{i},\dot{{q}}_{i})\in\mathcal{Q}\times\Real^{n_{q}}$. For future purposes, the vector of GRFs for the agent $i$ is shown by $f_{i}:=\col\{f_{i}^{\ell}\,|\,\ell\in\mathcal{C}_{i}\}$. 

For the whole-body motion control of each agent, we develop a QP-based nonlinear controller that maps the desired optimal trajectories and GRFs, generated by the high-level MPC, to the full-order model. For this purpose, we consider the following time-varying and holonomic output functions, referred to as \textit{virtual constraints} \cite{Randy_Paper_LCSS}, to be regulated:
\begin{equation}\label{virtual_constraints}
    y_{i}(t,z_{i}):=y_{a}(q_{i}) - y_{\des}(t),
\end{equation}
where $y_{a}(q_{i})$ represents a set of controlled variables and $y_{\des}(t)$ denotes the desired evolution of the controlled variables in terms of a time-based phasing variable. In this paper, the controlled variables include the Cartesian coordinates of the robot's COM, the absolute orientation of the robot's body (i.e., Euler angles), and Cartesian coordinates of the swing leg ends. The desired evolution of the COM position and Euler angles are generated by the high-level MPC. In addition, the desired evolution of the swing leg ends' coordinates are taken as a B\'ezier polynomial connecting the current footholds to the upcoming ones, computed based on Raibert’s heuristic \cite[Eq. (2.4), pp. 46]{raibert1986legged}. 

We next implement the standard I-O linearization technique \cite{Isidori_Book} to differentiate the local outputs \eqref{virtual_constraints} twice along the full-order dynamics \eqref{full_order_model} while ignoring the interaction forces between the agents. This results in the following output dynamics
\begin{equation}\label{output_dyn}
    \ddot{y}_{i}=\Phi_{\tau}(z_{i})\,\tau_{i} + \Phi_{f}(z_{i})\,f_{i} + \phi(z_{i}) = -K_{P}\,y_{i} -K_{D}\,\dot{y}_{i} + \delta_{i},
\end{equation}
where $\Phi_{\tau}(z_{i})$, $\Phi_{f}(z_{i})$, and $\phi(z_{i})$ are proper matrices and vectors computed based on I-O linearization and Lie derivatives similar to \cite[Appendix A]{Jeeseop_Hamed_ASME}. Moreover, $K_{P}$ and $K_{D}$ are positive definite matrices, and $\delta_{i}$ is a slack variable to be used later for the feasibility of the QP-based nonlinear controller. Unlike the high-level trajectory planner of Section \ref{Sec:Predictive Control based Planner} that takes into account the interaction terms, the low-level nonlinear controller ignores the interaction forces. In particular, our numerical results in Section \ref{Sec:Numerical and Experimental Validations} suggest that considering holonomic constraints for trajectory planning is crucial for stabilizing cooperative locomotion. However, the optimal trajectories, generated by the high-level MPC, can be robustly mapped to the full-order model without considering the interaction terms at the low level. This reduces the complexity of the distributed and full-order model controllers.  

By stacking together the Cartesian coordinates of the stance leg ends and then differentiating them twice, one can get an additional constraint to express zero acceleration for the stance leg ends. In particular, we have
\begin{equation}\label{contact_eq}
    \ddot{r}_{\textrm{foot},i}=\Theta_{\tau}(z_{i})\,\tau_{i}+\Theta_{f}(z_{i})\,f+\theta(z_{i})=0,
\end{equation}
where $r_{\textrm{foot},i}:=\col\{r_{\textrm{foot},i}^{\ell}\,|\,\ell\in\mathcal{C}_{i}\}$ is a vector containing the Cartesian coordinates of the stance feet for the agent $i$. Moreover, $\Theta_{\tau}(z_{i})$, $\Theta_{f}(z_{i})$, and $\theta(z_{i})$ are proper matrices and vectors computed based on I-O linearization. We them employ the following real-time and strictly convex QP \cite{Randy_Paper_LCSS} to solve for feasible $(\tau_{i},f_{i},\delta_{i})$ at $1$kHz to meet the output dynamics \eqref{local_dyn} and the contact equation \eqref{contact_eq}
\begin{alignat}{4}
&\!\min_{(\tau_{i},f_{i},\delta_{i})} && \frac{\gamma_{1}}{2}\|\tau_{i}\|^{2}+\frac{\gamma_{2}}{2}\|f_{i}-f_{\textrm{des},i}\|^{2}+\frac{\gamma_{3}}{2}\|\delta_{i}\|^{2}\nonumber\\
&\quad\textrm{s.t.}&&\!\! \Phi_{\tau}(z_{i})\,\tau_{i}+\Phi_{f}(z_{i})\,f_{i}+\phi(z_{i})=-K_{P}\,y_{i}-K_{D}\,\dot{y}_{i}+\delta_{i}\nonumber\\
& && \!\!\Theta_{\tau}(z_{i})\,\tau_{i}+\Theta_{f}(z_{i})\,f_{i}+\theta(z_{i})=0\nonumber\\
& && \!\!\tau_{i}\in\mathcal{T},\quad f_{i}^{\ell}\in\mathcal{FC},\quad\,\,\,\forall \ell\in\mathcal{C}_{i},\label{qp_HZD_controller}
\end{alignat}
where $\gamma_{1}$, $\gamma_{2}$, and $\gamma_{3}$ are positive weighting factors, and $f_{\des,i}$ represents the desired evolution of the GRFs generated by the high-level MPC. 
The cost function of \eqref{qp_HZD_controller} tries to minimize the effect of the slack variable $\delta_{i}$ in the output dynamics \eqref{output_dyn} via a high weighting factor $\gamma_{3}$ while 1) imposing the actual GRFs of the full-order model $f_{i}$ to follow the prescribed force profile $f_{\des,i}$ with the weighting factor $\gamma_{2}$, and 2) having the minimum-power torques with the weighting factor $\gamma_{1}$. 


\begin{figure}[t!]
\centering
\includegraphics[draft=false, width=\linewidth]{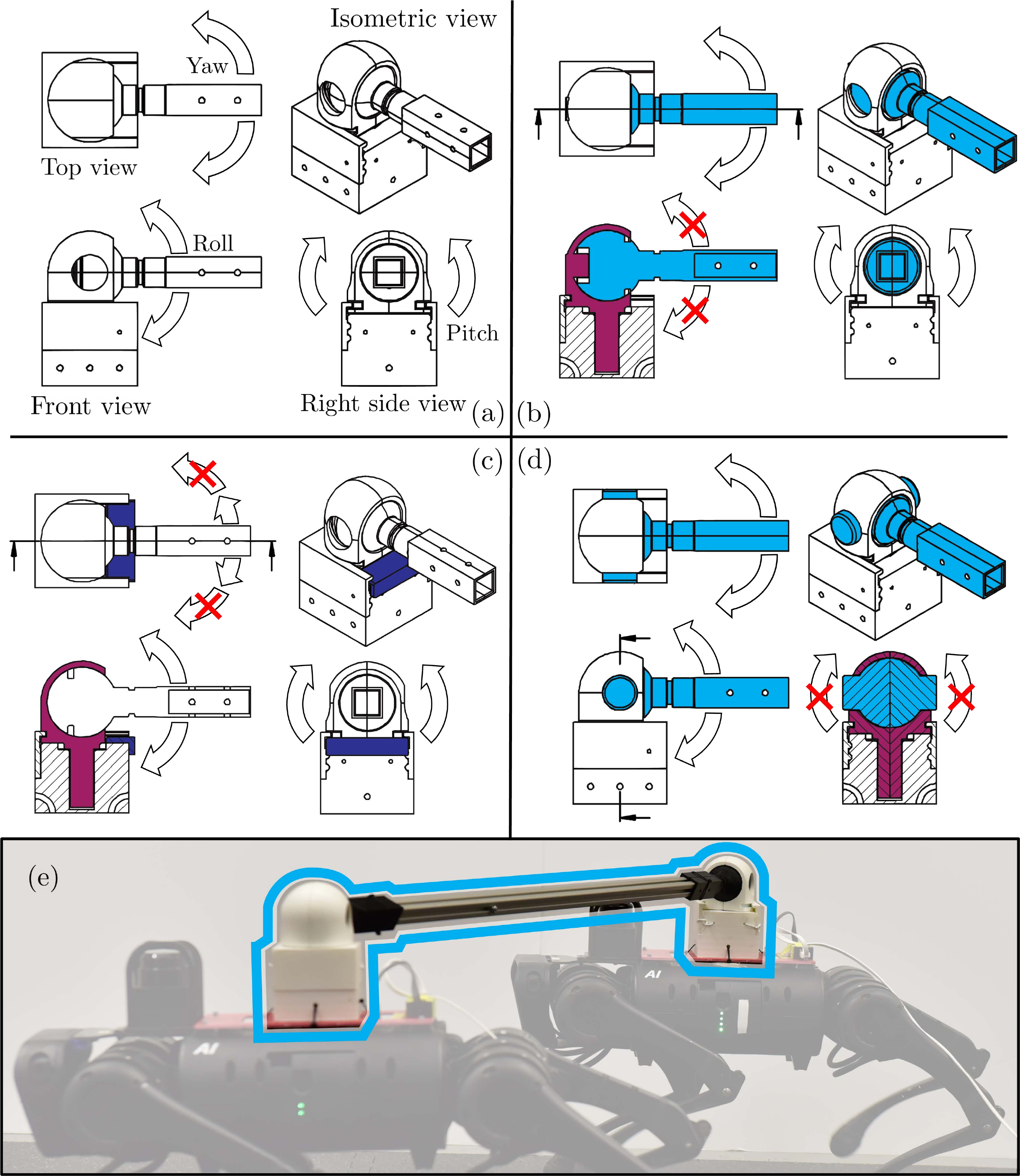}
\caption{Illustration of the mechanisms designed to be mounted on the torso of each robot to holonimcally constrain the motion of agents. The mechanism in (a) can implement the holonomic constraint in \eqref{holonomic_constraint} with free ball joints. The mechanisms in (b), (c), and (d) implement the constraint \eqref{holonomic_constraint} while also restricting the roll, yaw, and pitch motions, respectively. The mechanism implemented on top of the robots is illustrated in (e).}
\label{fig:dofregulation}
\vspace{-0.5em}
\end{figure}

\section{Numerical and Experimental Validations}
\label{Sec:Numerical and Experimental Validations}

This section aims to validate the proposed layered control architecture composed of the high-level centralized and distributed MPC algorithms and the low-level distributed nonlinear controllers via extensive numerical simulations and experiments. We study both the reduced- and full-order models of cooperative locomotion in numerical simulations to show the robust stability of the collaborative gaits. We further experimentally investigate the robustness of the trajectories with a team of two holonomically constrained A1 robots, as shown in Fig. \ref{fig:2agents_snap}.

\vspace{-0.5em}
\subsection{Closed-Loop System}
\label{sec:Closed-Loop System}

\subsubsection{\textbf{Robot hardware and gait}}
The hardware platform considered here, the A1 robot, is a torque-controlled quadrupedal robot platform with 18 DOFs and 12 actuators. More specifically, 12 DOFs of the system represent the actuated DOFs of the legs' joints. Each leg consists of a 2-DOF hip joint (roll and pitch) and a 1-DOF knee joint (knee pitch). The remaining 6 DOFs describe the unactuated position and orientation of the body with respect to the inertial world frame. The robot is approximately 12.45 (kg) and stands up to about 0.3 (m) off the ground. This work considers a standing height of 0.26 (m) for all experiments. Here, the position of the interactions points with respect to COMs in the body frames $\{B_{i}\}$ is taken as $d_{i}^{B_{i}}=\textrm{col}(0,0,0.15)$ (m) for all $i \in \mathcal{I}$ (see \eqref{interaction_points}). Different mechanisms are designed to holonomically constrain the motion of two robots with ball joints and an adjustable bar length between the agents (see Fig. \ref{fig:dofregulation}). Furthermore, the mechanisms can limit the ball joints to add further constraints on their Euler angles. For numerical and experimental studies in Sections \ref{Sec:Numerical Validation} and \ref{Sec:Experimental Validation and Robustness Analysis}, the nominal length of the bar is 1 (m) (see Fig. \ref{fig:2agents_snap}). However, we can alter it to 0.75 (m) and 1.5 (m) for the robustness analysis.

\begin{figure}[t!]
\centering
\includegraphics[draft=false, width=\linewidth]{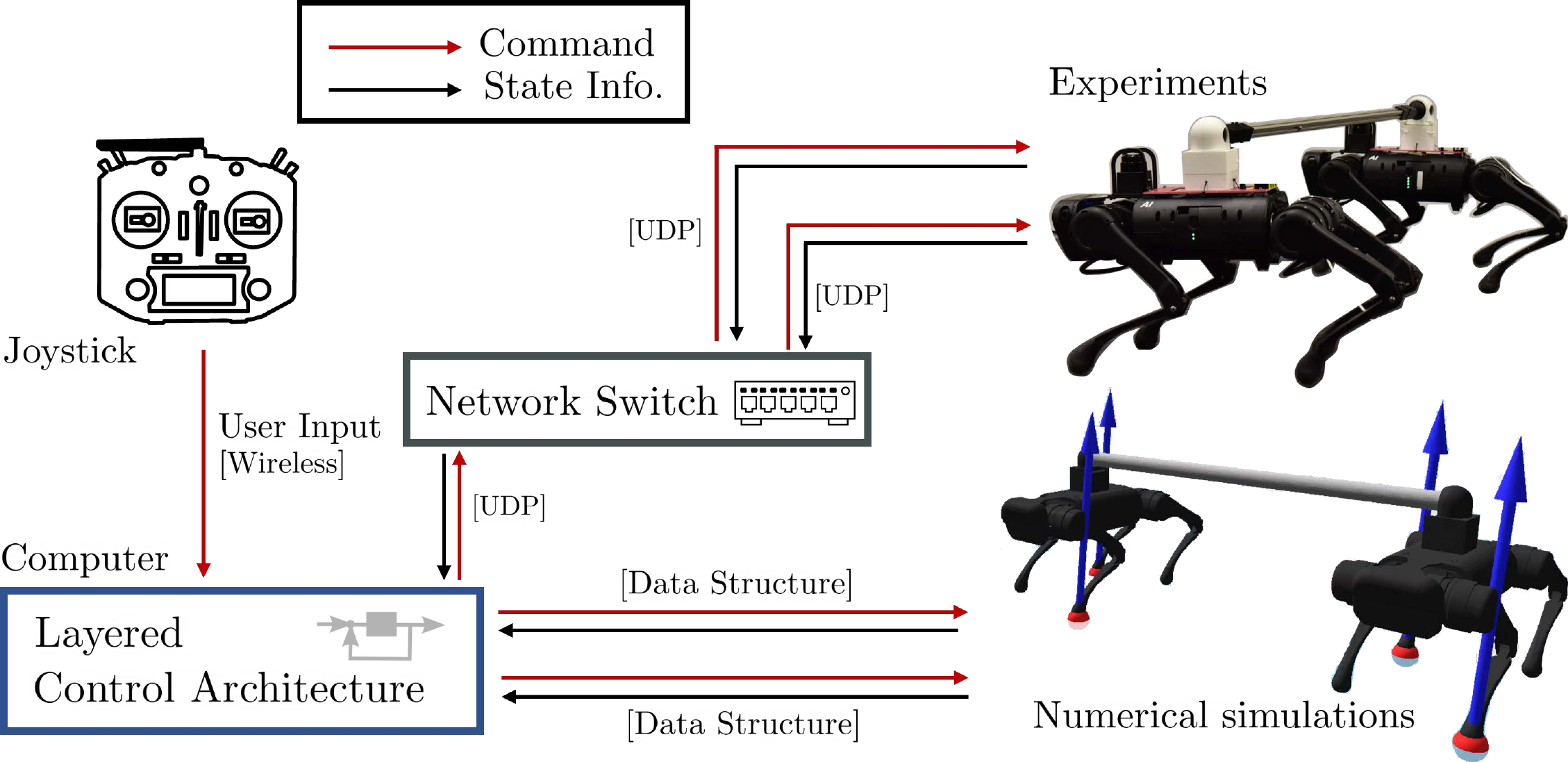}
\caption{Numerical and experimental validation system setup. Here, the joystick commands the desired velocity trajectories to the trajectory planner of each agent. Both agents are controlled by one joystick. Joystick sends out the desired trajectories on both numerical simulations and experimental validations. The network switch is used to build the connection between the computer and two agents without IP address collision. UDP communication protocol through Ethernet cables is used in experimental validations.}
\label{fig:connectionmap}
\vspace{-0.5em}
\end{figure}

In the following sections, we study a cooperative trot gait with a swing time of 0.2 (s) and at different speeds up to 0.5 (m/s) and subject to external disturbances, uncertainties in holonomic constraints, unknown payloads up to $55\%$ uncertainty in one robot's mass, and on different terrains (e.g., slippery surfaces, wooden blocks, gravel, mulch, and grass).  

\subsubsection{\textbf{Computation, control loop, and network}}
We use RaiSim \cite{RAISIM} to simulate both the interconnected reduced- and full-order models numerically. The proposed high-level centralized and distributed MPC algorithms for trajectory planning and the low-level distributed nonlinear controllers for whole-body motion control are solved using qpSWIFT \cite{qpSWIFT} at 200 Hz and 1 kHz, respectively. A joystick is used to command the desired velocity trajectories to the high-level trajectory planner. The joystick includes two 2-DOFs gimbals, six auxiliary switches, and two knobs for the controlling purpose (see Fig. \ref{fig:connectionmap}). The gimbals are used to command the desired speed, whereas the switches allow us to simultaneously control both agents or individually command them. This control scheme allows us to coordinate the agents during cooperative locomotion and unexpected scenarios effectively. This will be discussed further in Section \ref{subsec:Limitations and Future Study}. Moreover, we remark that the joystick commands the desired trajectories for both the numerical simulations and experimental validations. The joystick connects with the computer through a 2.4 GHz wireless channel as described in Fig. \ref{fig:connectionmap}.

The proposed layered controller, including the MPC-based trajectory planners and distributed nonlinear controllers, is solved on an off-board laptop computer with an i7-10750H CPU running at 2.60 GHz and 16 GB RAM. For the experiment, we use a network switch in the connection between the robotic team and the computer. The connection diagram is illustrated in Fig. \ref{fig:connectionmap}. The switch supports 1000 Mbps gigabit Ethernet with five ports. The robot IP addresses are redefined to avoid IP collision during communication. Here, we also define the IP routing table and proper IP address on the computer to communicate with both agents without data packet confusion. Internally, a UDP protocol through Ethernet cables is used to communicate between the computer and the robots. The data structure in C++ is used for numerical simulations to communicate between the layered control architecture and the simulation environment.

\begin{figure}[t!]
\centering
\includegraphics[draft=false, width=\linewidth]{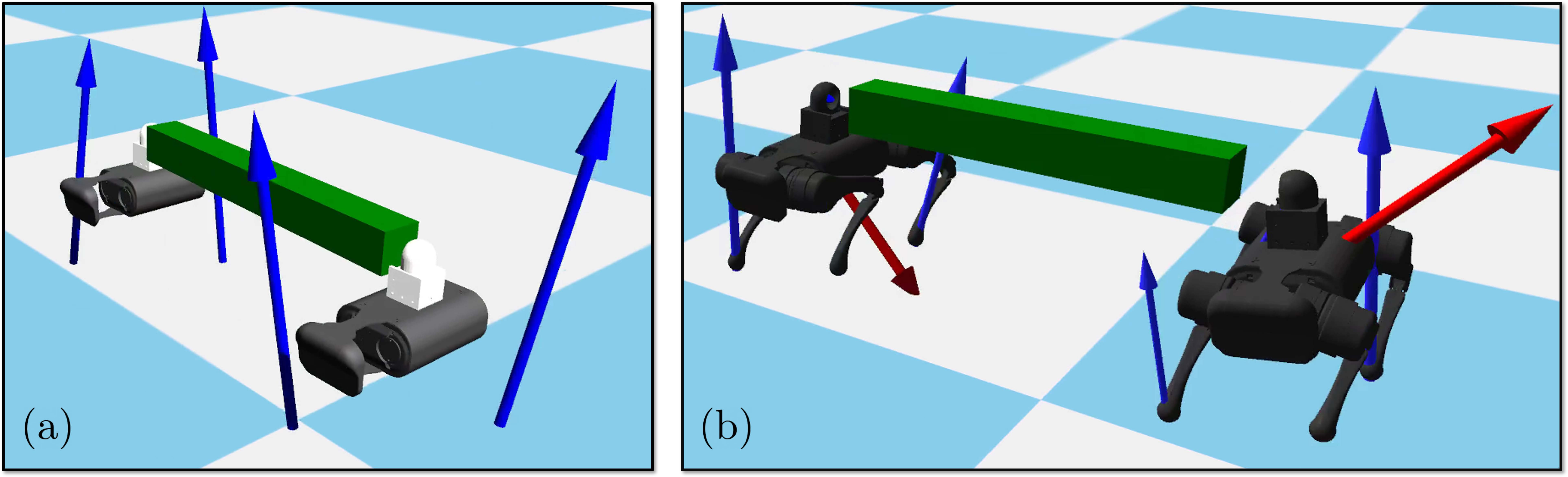}
\vspace{-1.5em}
\caption{Snapshots demonstrating the performance of the proposed control approach in numerical simulations. The left plot shows the snapshot of the numerical simulation with the interconnected reduced-order model (torso dynamics) and subject to a 5 (kg) payload ($40\%$ uncertainty in one robot's mass) between the agents. The right plot shows the snapshot of the numerical simulation with the full-order model and subject to a 5 (kg) payload between agents and unknown time-varying external disturbances applied to the robots. Arrows at the leg ends describe the GRFs, and the ones on the torso represent the external disturbance forces. The payload is illustrated with a box.} 
\vspace{-0.5em}
\label{fig:numerical_snap}
\end{figure}

\subsubsection{\textbf{Tuning controllers}}
The control horizon for both the centralized and distributed MPC is taken as $N=5$ discrete-time samples, where the time discretization at the high level is 5 (ms). The centralized and distributed MPC algorithms in \eqref{centralized_MPC} and \eqref{distributed_MPC} have 245 and 125 decision variables, respectively. The stage cost gain of the centralized MPC is tuned as $\bm{Q} = \mathrm{diag}\{Q_{rc1} \ Q_{rc2} \ Q_{\dot{r}c1} \ Q_{\dot{r}c2} \ Q_{\xi 1} \ Q_{\xi 2} \ Q_{\omega1} \ Q_{\omega2}\} \in \Real^{24\times 24}$, where $\bm{Q}_{rci} = 10^{5}\times\textnormal{diag}\{3 \ 300 \ 30\}$, $\bm{Q}_{\dot{r}ci} = 10^4\,\iden_{3\times 3}$, $\bm{Q}_{\xi i} = 10^8\,\iden_{3\times 3}$, and $\bm{Q}_{\omega i} = 5\times 10^3\,\iden_{3\times 3}$, $i\in\mathcal{I}$. The terminal cost gain of the centralized MPC is also tuned as $\bm{P} = 10^{-1} \bm{Q} \in \Real^{24\times 24}$. The input gains of the centralized MPC are chosen as $\bm{R}_{u} = 10^{-2}\,\iden_{24\times 24}$ and $\bm{R}_{\lambda} = 10^{4}$. In a similar manner, the stage cost gain and terminal cost gain of the distributed MPC on the $i$-th agent are tuned as $\bm{Q}_i = \mathrm{diag}\{Q_{rci} \ Q_{\dot{r}ci} \ Q_{\xi i} \ Q_{\omega i}\} \in \Real^{12\times 12}$ and $\bm{P}_i = 10^{-1} \bm{Q}_i \in \Real^{12\times 12}$. The input gains of the distributed MPC are finally chosen as $\bm{R}_{u} = 10^{-2}\,\iden_{12\times 12}$ and $\bm{R}_{\lambda} = 10^{4}$. Additionally, we choose the weighting factor for the agreement protocol in \eqref{distributed_MPC} as $w = 10$, and the averaging factors in \eqref{distributed_MPC} are chosen as $a_{ii} = a_{ij} = 0.5$ for all $i\neq j \in\mathcal{I}$. The friction coefficient for both the centralized MPC and distributed MPC algorithms is assumed to be $\mu = 0.6$. However, the experiments on slippery surfaces assume a lower friction coefficient of $\mu = 0.3$. For the low-level and distributed nonlinear controllers in \eqref{qp_HZD_controller}, the weighting factors for the joint-level torques, force tracking error, and slack variables are chosen as $\gamma_1 = 10^2$, $\gamma_2 = 10^4$, and $\gamma_3 = 10^6$, respectively. We finally remark that the low-level controller uses the same friction coefficient values from the high-level MPC.

The computation time of the centralized and distributed MPC algorithms under nominal conditions is approximately 1.38 (ms) and 0.41 (ms), respectively. This shows that the solve time with the proposed distributed MPC is reduced by $70\%$. Furthermore, the computation time of the low-level distributed nonlinear controllers is about 0.12 (ms). 

\begin{figure}[t!]
\centering
\includegraphics[draft=false, width=\linewidth]{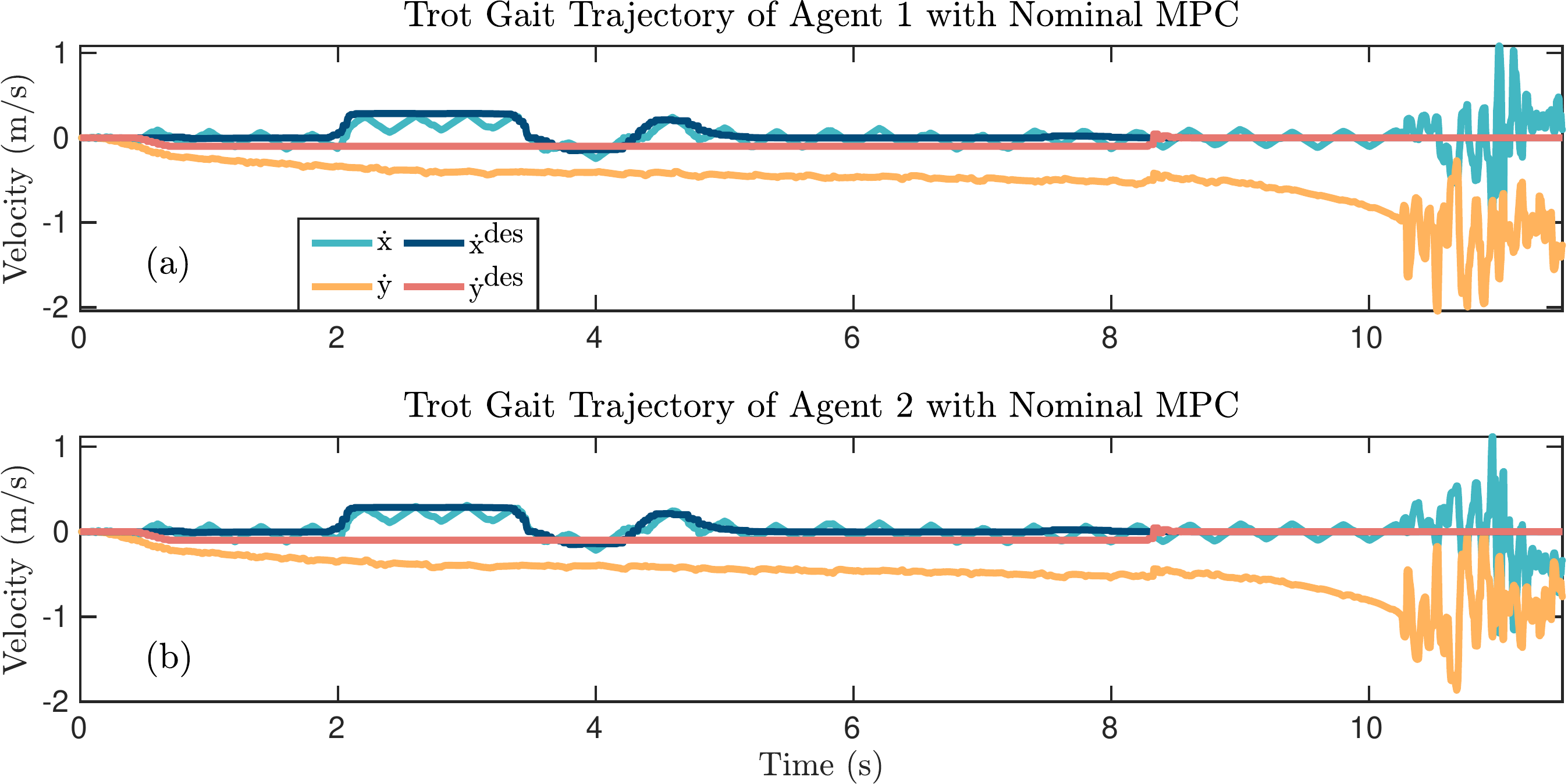}
\vspace{-1.5em}
\caption{Plots of the desired and actual velocities of the closed-loop interconnected reduced-order model for two agents in the numerical simulation. Here, two nominal MPCs are applied to the reduced-order models of agents to generate optimal GRFs \textit{without} considering the holonomic constraints between them. The joystick provides desired trajectories. The instability of the cooperative gait is evident.}
\label{fig:numerical_reduced_order_nominalMPC}
\vspace{-0.5em}
\end{figure}

\vspace{-0.5em}
\subsection{Numerical Validation}
\label{Sec:Numerical Validation}

\subsubsection{\textbf{Simulation with the reduced-order model}} 

\begin{figure}[t!]
\centering
\includegraphics[draft=false, width=\linewidth]{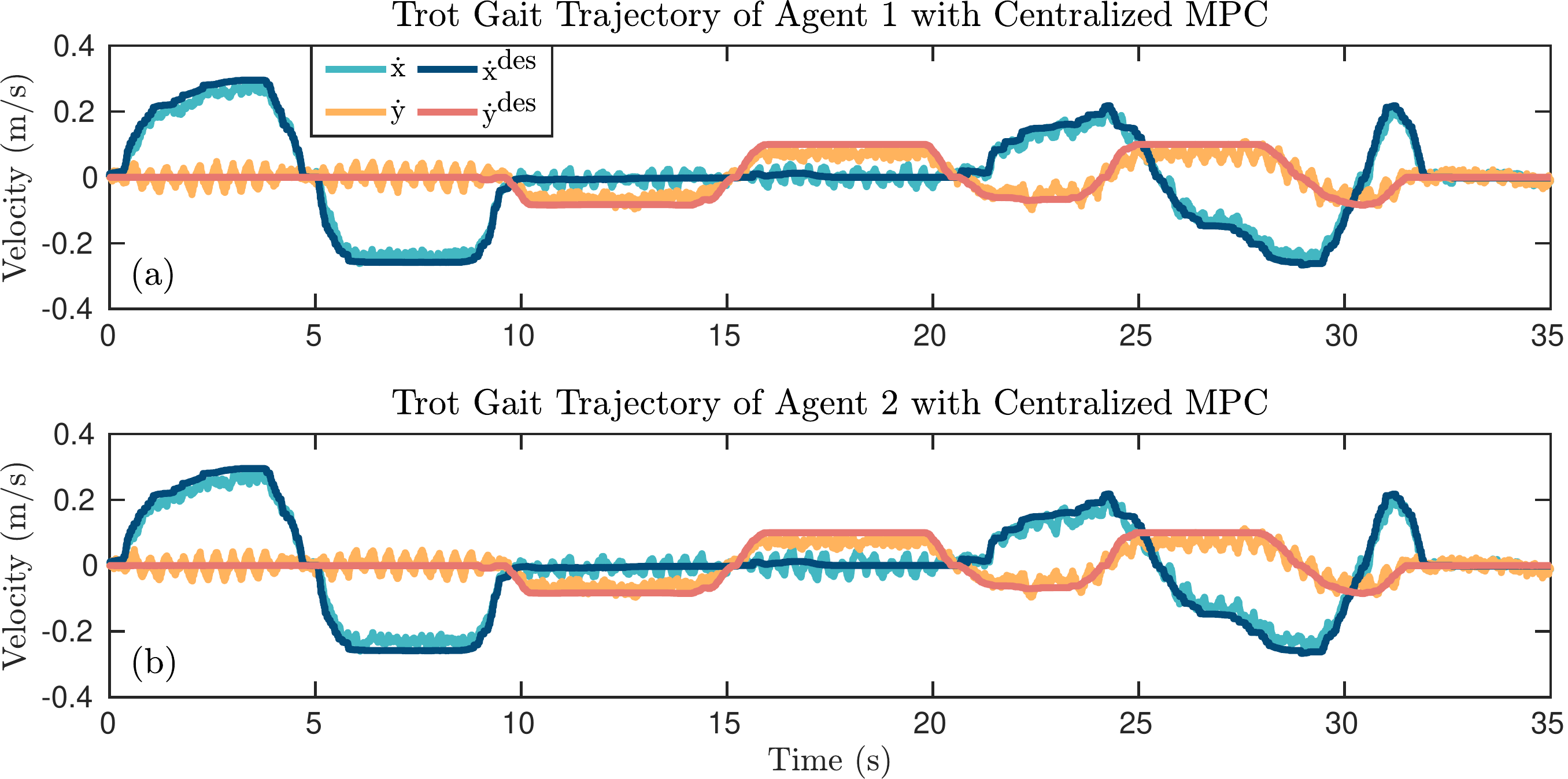}
\vspace{-1.5em}
\caption{Plots of  the desired and actual velocities of the closed-loop interconnected reduced-order model for two agents in the numerical simulation. 
Here, the optimal GRFs are generated by the centralized MPC \eqref{centralized_MPC} and are applied to the reduced-order models subject to an unknown payload of 5 (kg) between agents. The joystick provides desired trajectories. The robust tracking is evident.}
\label{fig:numerical_reduced_order_centralizedMPC}
\vspace{-0.5em}
\end{figure}

\begin{figure}[t!]
\centering
\includegraphics[draft=false, width=\linewidth]{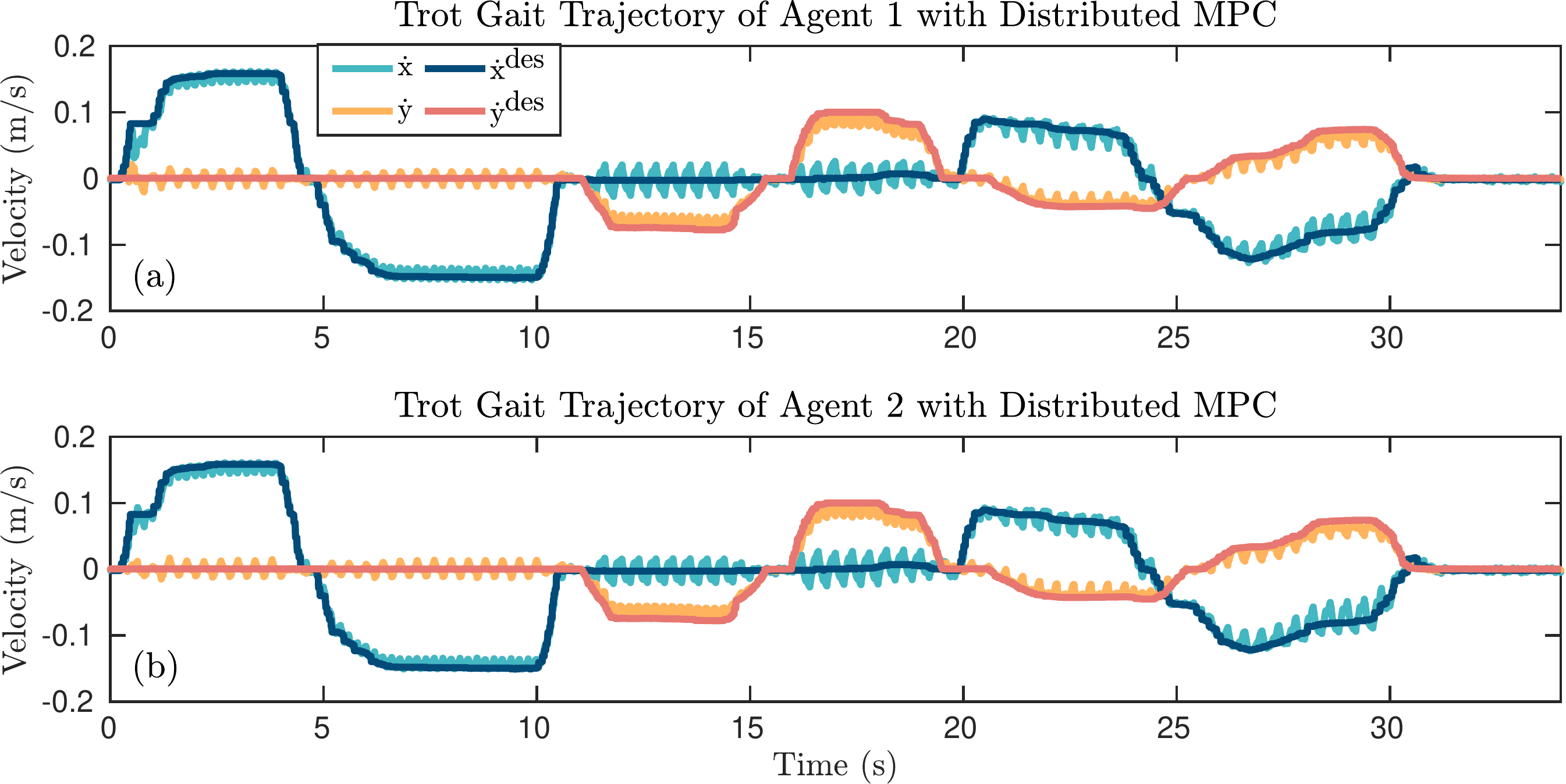}
\vspace{-1.5em}
\caption{Plots of the desired and actual velocities of the closed-loop interconnected reduced-order model for two agents in the numerical simulation. 
Here, the optimal GRFs are generated by the distributed MPCs \eqref{distributed_MPC} and are applied to the reduced-order models subject to an unknown payload of 5 (kg) between agents. The joystick provides desired trajectories. The robust tracking is evident.}
\label{fig:numerical_reduced_order_distributedMPC}
\vspace{-0.5em}
\end{figure}

We model the interconnected SRB dynamics in the RaiSim environment for numerical validation and apply the optimal GRFs generated from the proposed centralized \eqref{centralized_MPC} and distributed MPC \eqref{distributed_MPC} algorithms. In addition, for comparison purposes, we apply the GRFs generated from the nominal MPC that considers a standard SRB model \textit{without} the holonomic constraints to this interconnected model. An overview of the numerical simulation environment for the interconnected reduced-order model is illustrated in Fig. \ref{fig:numerical_snap}(a). 
The evolution of the desired and actual COM velocities using the nominal MPC is depicted in Fig. \ref{fig:numerical_reduced_order_nominalMPC}. It is evident that the nominal MPC, which does not consider the holonomic constraint between agents, \textit{cannot} stabilize the interconnected reduced-order system. On the other hand, the interconnected SRB model performs stable cooperative locomotion when integrated with the GRFs generated from the proposed centralized and distributed MPCs as shown in Figs. \ref{fig:numerical_reduced_order_centralizedMPC} and \ref{fig:numerical_reduced_order_distributedMPC}, respectively. 
In these simulations, an unknown payload of 5 (kg) ($40\%$ uncertainty in one robot's mass) is considered between the agents (i.e., in the middle of the bar), and the joystick provides the desired trajectories. Figures \ref{fig:numerical_reduced_order_centralizedMPC} and \ref{fig:numerical_reduced_order_distributedMPC} illustrate that the closed-loop interconnected reduced-order model robustly tracks the time-varying desired trajectories subject to unknown payloads. Animations of all simulations can be found online \cite{YouTube_CoopertaiveLoco}. 

\subsubsection{\textbf{Simulation with the full-order model}}
\begin{figure}[t!]
\centering
\includegraphics[draft=false, width=\linewidth]{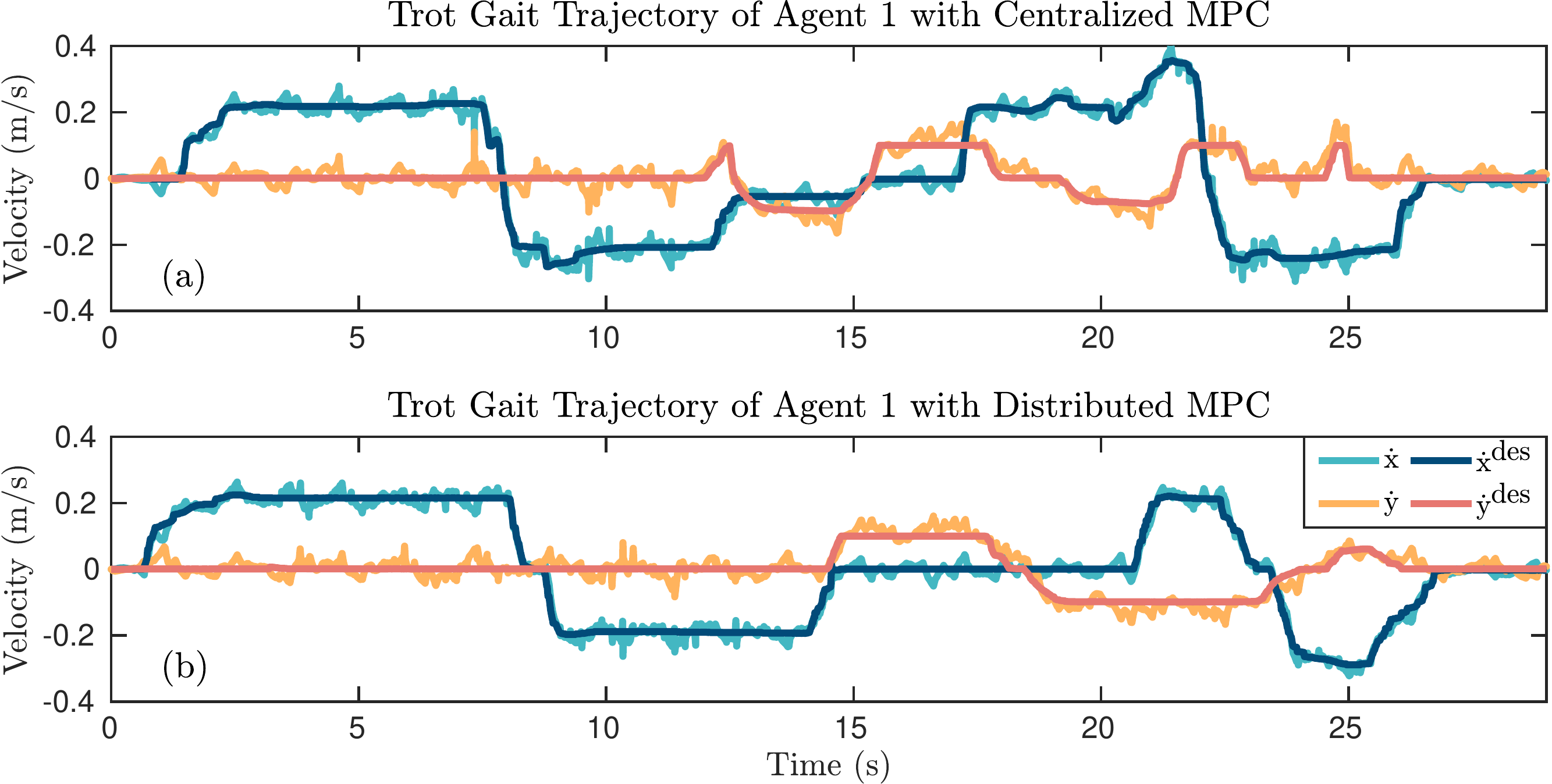}
\vspace{-1.5em}
\caption{Comparison between the desired velocities and optimal velocities, generated with the high-level centralized and distributed MPCs, for the closed-loop interconnected full-order model in RaiSim. The joystick provides time-varying reference trajectories. The figure depicts the optimal trajectories generated by (a) the centralized MPC \eqref{centralized_MPC} and (b) the distributed MPC \eqref{distributed_MPC} for agent $1$. Here, we consider a trot gait over rough terrain with an unknown payload of 5 (kg) between the agents and subject to unknown, time-varying, and external disturbance forces applied to the robots.}
\label{fig:numerical_full_order_trot_w_disturbances}
\vspace{-0.5em}
\end{figure}

We next numerically study the performance of the closed-loop system with the interconnected full-order dynamical model in RaiSim. Here, the proposed layered control approach is employed, including the centralized and distributed MPC algorithms for trajectory planning and nonlinear controllers for whole-body motion control. The desired time-varying trajectories are generated using the joystick. The high-level MPC then generates optimal GRFs and reduced-order trajectories from the centralized and distributed algorithms. The distributed low-level controller computes the corresponding joint-level torques to impose the full-order model to track the optimal trajectories. An overview of the numerical simulation environment for the full-order model is illustrated in Fig. \ref{fig:numerical_snap}(b). The desired trajectories provided by the joystick together with the optimal trajectories computed by the centralized and distributed MPC are depicted  in Figs. \ref{fig:numerical_full_order_trot_w_disturbances}(a) and \ref{fig:numerical_full_order_trot_w_disturbances}(b). Due to the similarity of the plots for agents, Fig. \ref{fig:numerical_full_order_trot_w_disturbances} only includes the trajectories for agent 1. Here, we consider the trot gait over a randomly generated rough terrain with a maximum height of 5 (cm) ($19\%$ uncertainty in the robot's nominal height). The gait is also subject to an unknown payload of 5 (kg) and an unknown sinusoidal external disturbance force with the magnitude of 20 (N) and the period of 1.0 (s), 0.7 (s), and 0.4 (s) along the $x$-, $y$-, and $z$-directions, respectively. It is evident that the closed-loop system robustly tracks the desired trajectories. 


\begin{figure}[t!]
\centering
\includegraphics[width=\linewidth]{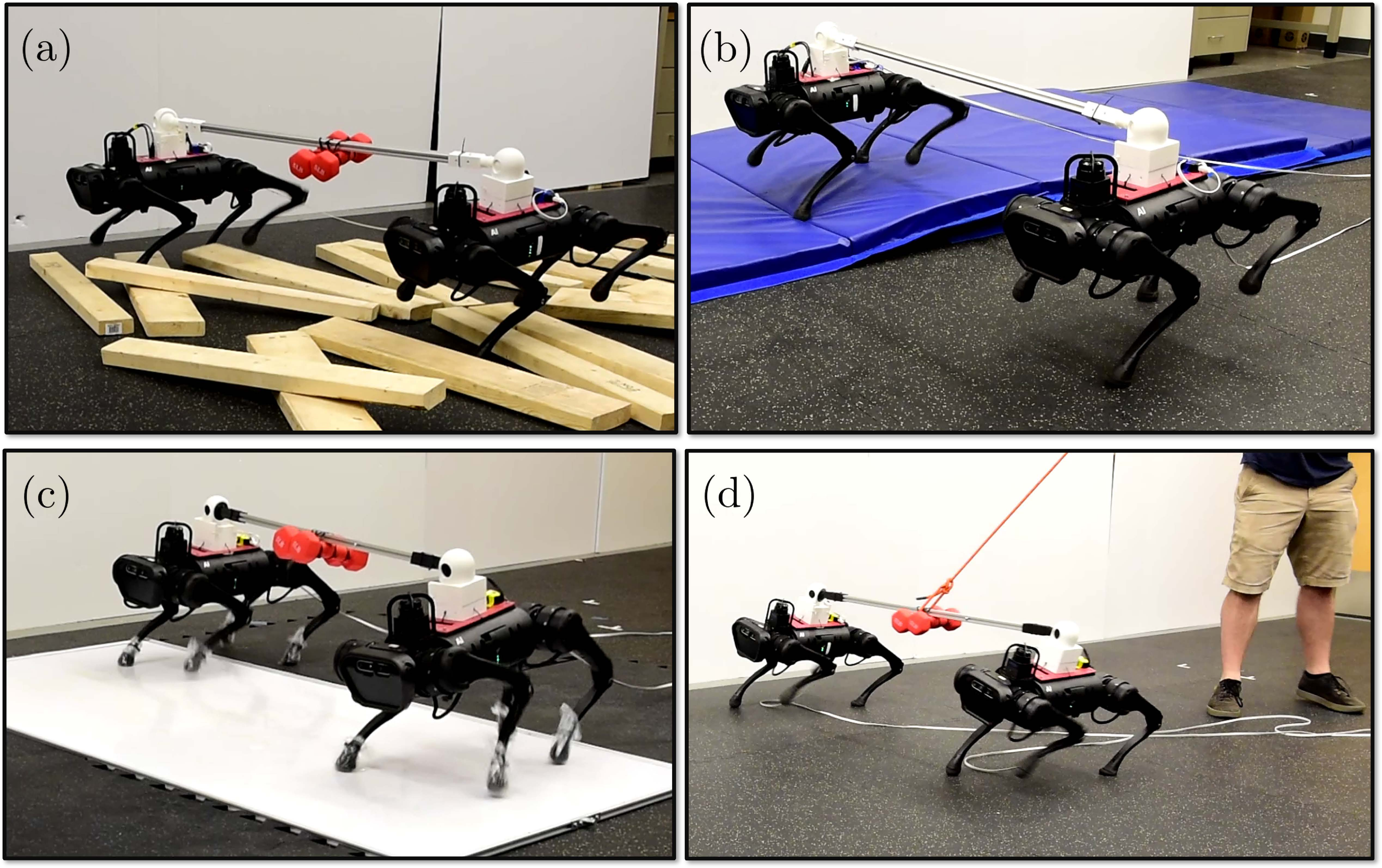}
\vspace{-1.5em}
\caption{Snapshots demonstrating the performance of the proposed layered control algorithm for a series of cooperative locomotion experiments. Indoor experiments: (a) rough terrain with the agents traversing arbitrarily displaced wooden blocks, (b) asymmetrical terrain with one agent being on a compliant surface and elevated by 10 (cm), (c) slippery surface covered by a cooking spray, and (d) tethered pulling. The robots are loaded with a payload of 4.53 (kg) ($36\%$ uncertainty in one robot's mass) in (a), (c), and (d). The friction coefficient is taken as $\mu=0.3$ in (c) and $\mu=0.6$ in (a), (b), and (d). Here, (a) and (b) show the snapshots where the centralized MPC is applied, while (c) and (d) show the snapshots where the distributed MPC is employed. Videos of all experiments are available online \cite{YouTube_CoopertaiveLoco}.} 
\vspace{-1.5em}
\label{fig:indoor_exp}
\end{figure}

\vspace{-0.5em}
\subsection{Experimental Validation and Robustness Analysis}
\label{Sec:Experimental Validation and Robustness Analysis}

This section experimentally validates the proposed layered control approach with the high-level centralized and distributed MPC algorithms and the low-level distributed nonlinear controllers. The robustness of the cooperative gaits on different indoor and outdoor terrains and subject to unknown payloads and external disturbances is evaluated. 

\subsubsection{\textbf{Indoor experiments with the centralized MPC}}
\begin{figure}[t!]
\centering
\includegraphics[draft=false, width=\linewidth]{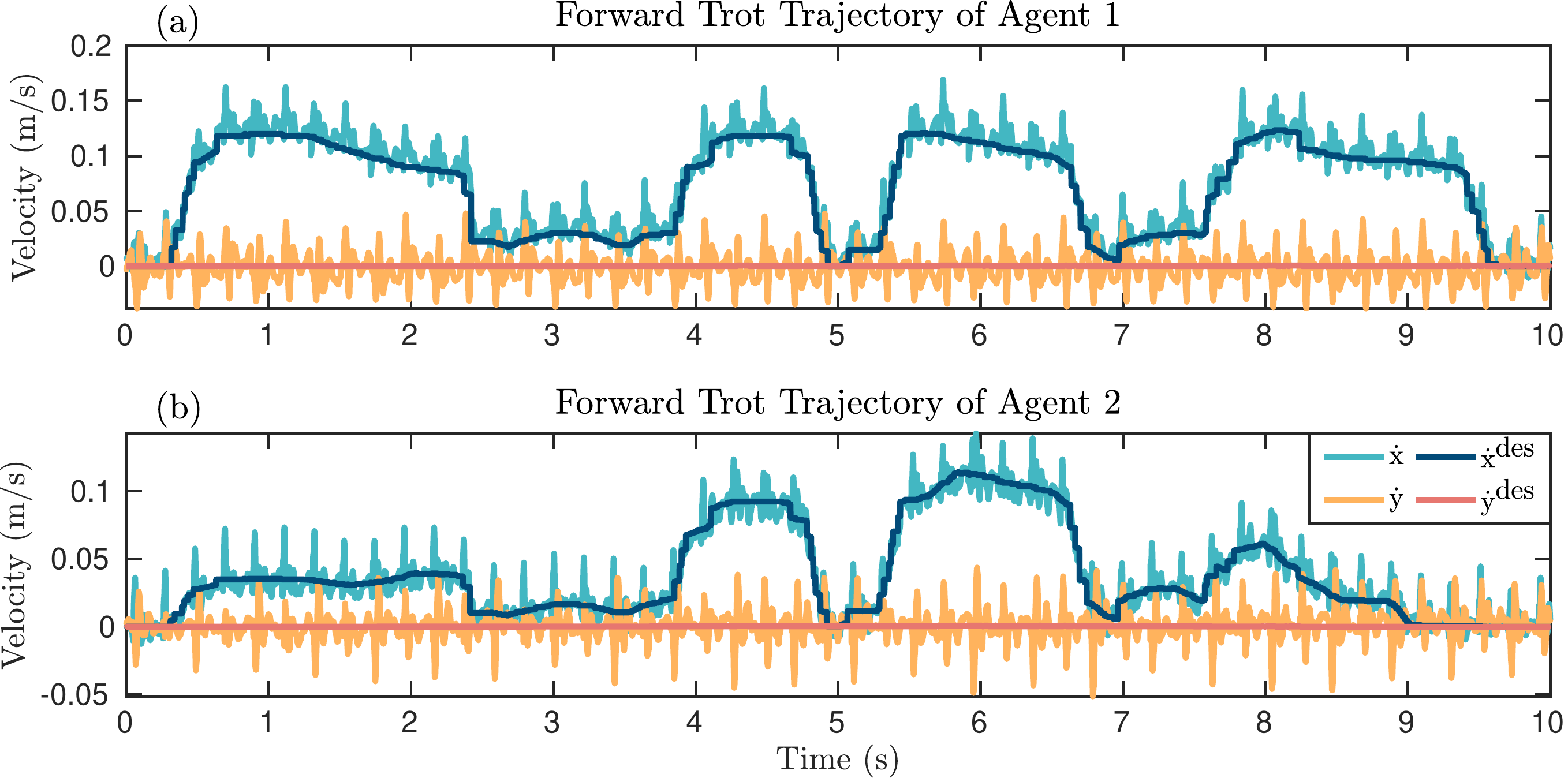}
\vspace{-1.5em}
\caption{Comparison between the desired and optimal velocities of the robots for the nominal trot experiment on flat ground. Optimal velocities are provided from the high-level centralized MPC. Time-varying desired trajectories are provided by the joystick to coordinate the robots' motions. It is observed that the centralized MPC's outputs can successfully track the desired trajectories.}
\label{fig:forwardtrot_centralized}
\vspace{-0.5em}
\end{figure}

\begin{figure}[t!]
\centering
\includegraphics[draft=false, width=\linewidth]{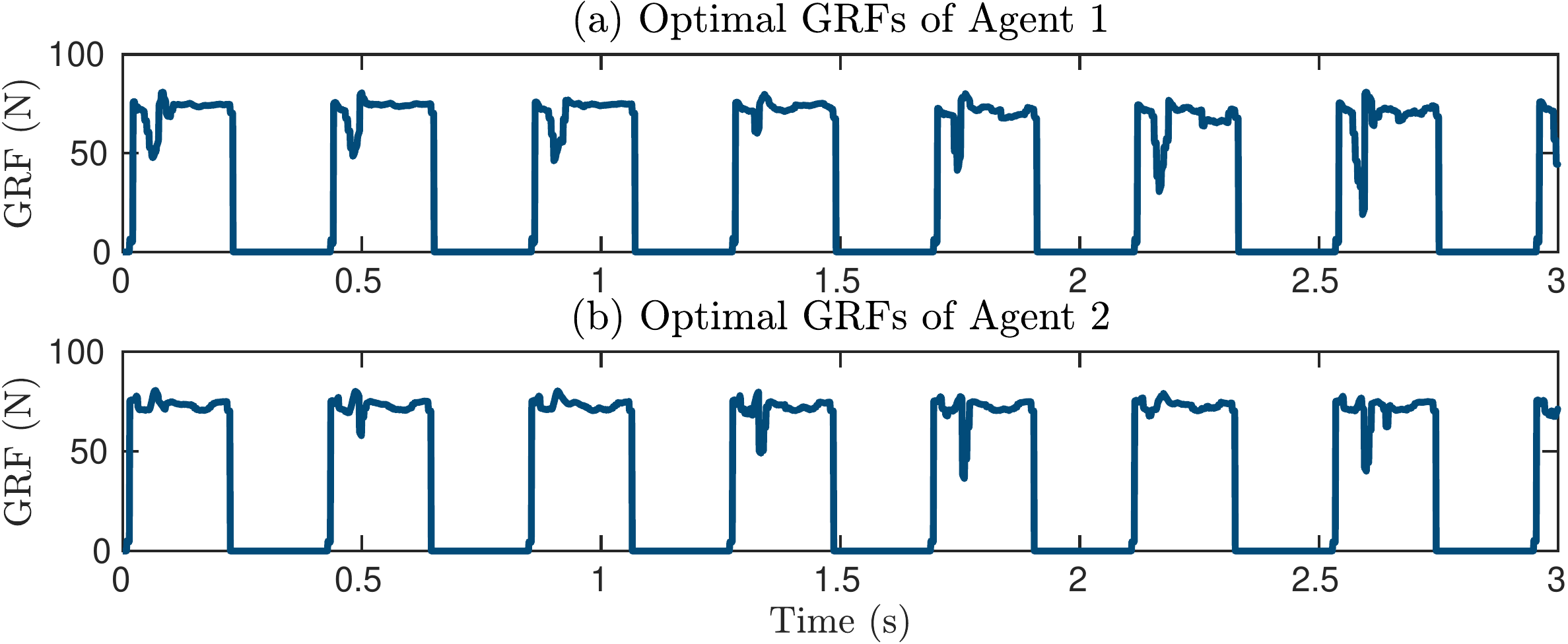}
\vspace{-1.5em}
\caption{Plots of the optimal GRFs generated from the centralized MPC for agents 1 and 2 during the nominal trot experiment on flat ground. The figure depicts the $z$ components of the optimal GRFs for the left front leg of each agent. We remark that the GRFs in the $z$-direction are close to 60 (N) since the trot gait is adopted and the total mass of the robot is 12.45 (kg).}
\label{fig:grf}
\vspace{-0.5em}
\end{figure}

In the indoor experiments, we employ the proposed layered control algorithm on two A1 robots subject to holonomic constraints, where ball joints are applied at the interaction points (see Fig. \ref{fig:indoor_exp}). We first investigate the nominal and cooperative trot gait with the centralized MPC algorithm on flat ground and without unknown disturbances. The desired and optimal COM trajectories, generated by the high-level MPC, together with the generated optimal GRFs, are illustrated in Fig. \ref{fig:forwardtrot_centralized} and Fig. \ref{fig:grf}, respectively. It is evident that the team of two A1 robots performs stable cooperative locomotion while the trajectory planner effectively tracks the time-varying desired trajectories. Furthermore, the optimal GRFs generated by the centralized MPC are feasible, with the vertical component value being close to 60 (N),  which is approximately the force required by each stance leg to support the total mass of each robot during trotting.

\begin{figure}[t!]
\centering
\includegraphics[draft=false, width=\linewidth]{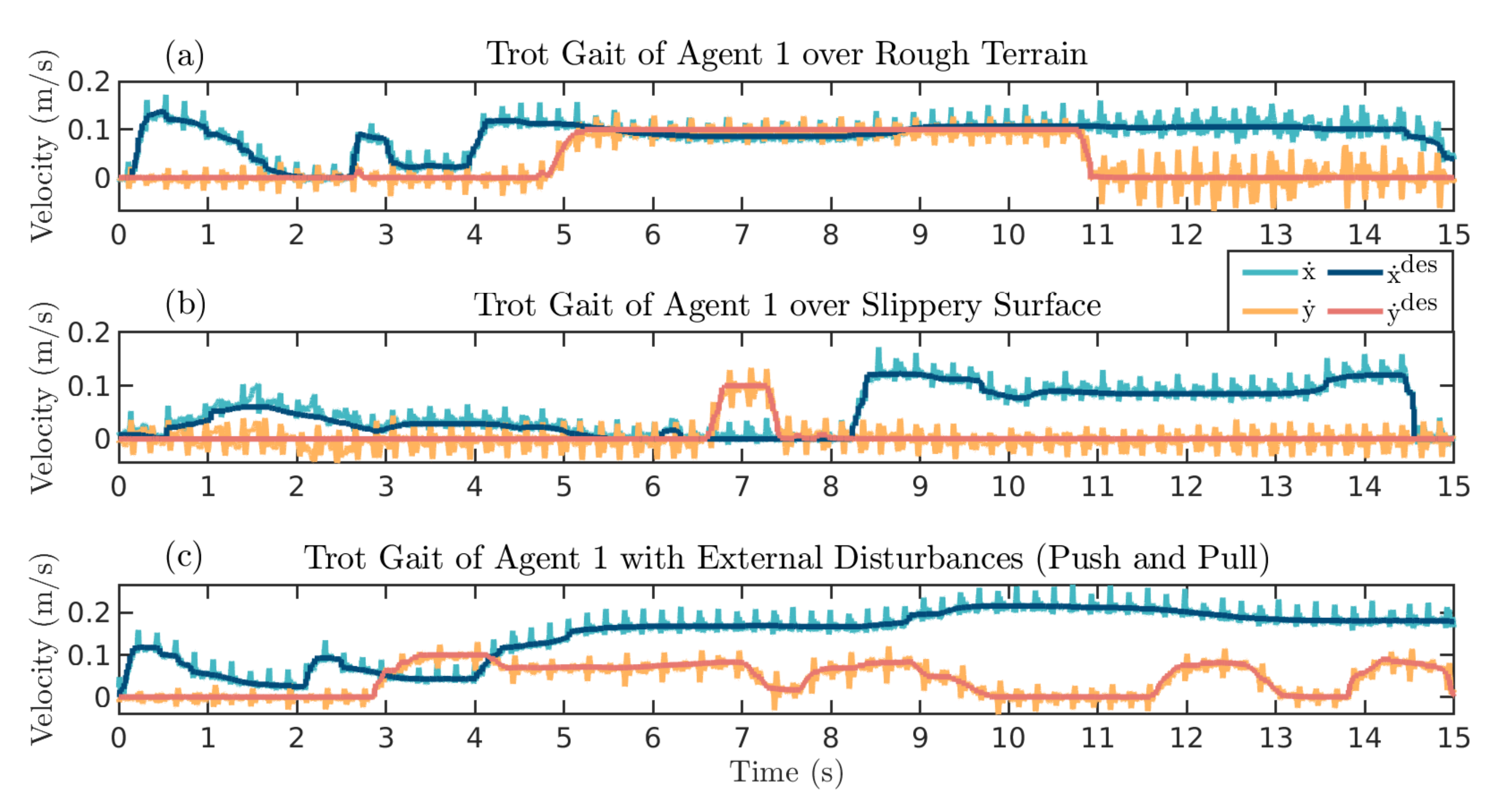}
\vspace{-1.5em}
\caption{Plots of the desired and optimal velocities for cooperative locomotion experiments on (a) rough terrain, (b) a slippery surface, and (c) subject to external disturbances with the centralized MPC. Time-varying desired trajectories are provided by the joystick. It is clear that the centralized MPC's outputs can robustly track the desired trajectories in the presence of uncertainties and disturbances.}
\label{fig:trot_vel_centralized_disturb}
\vspace{-0.5em}
\end{figure}

\begin{figure}[t!]
\centering
\includegraphics[draft=false, width=\linewidth]{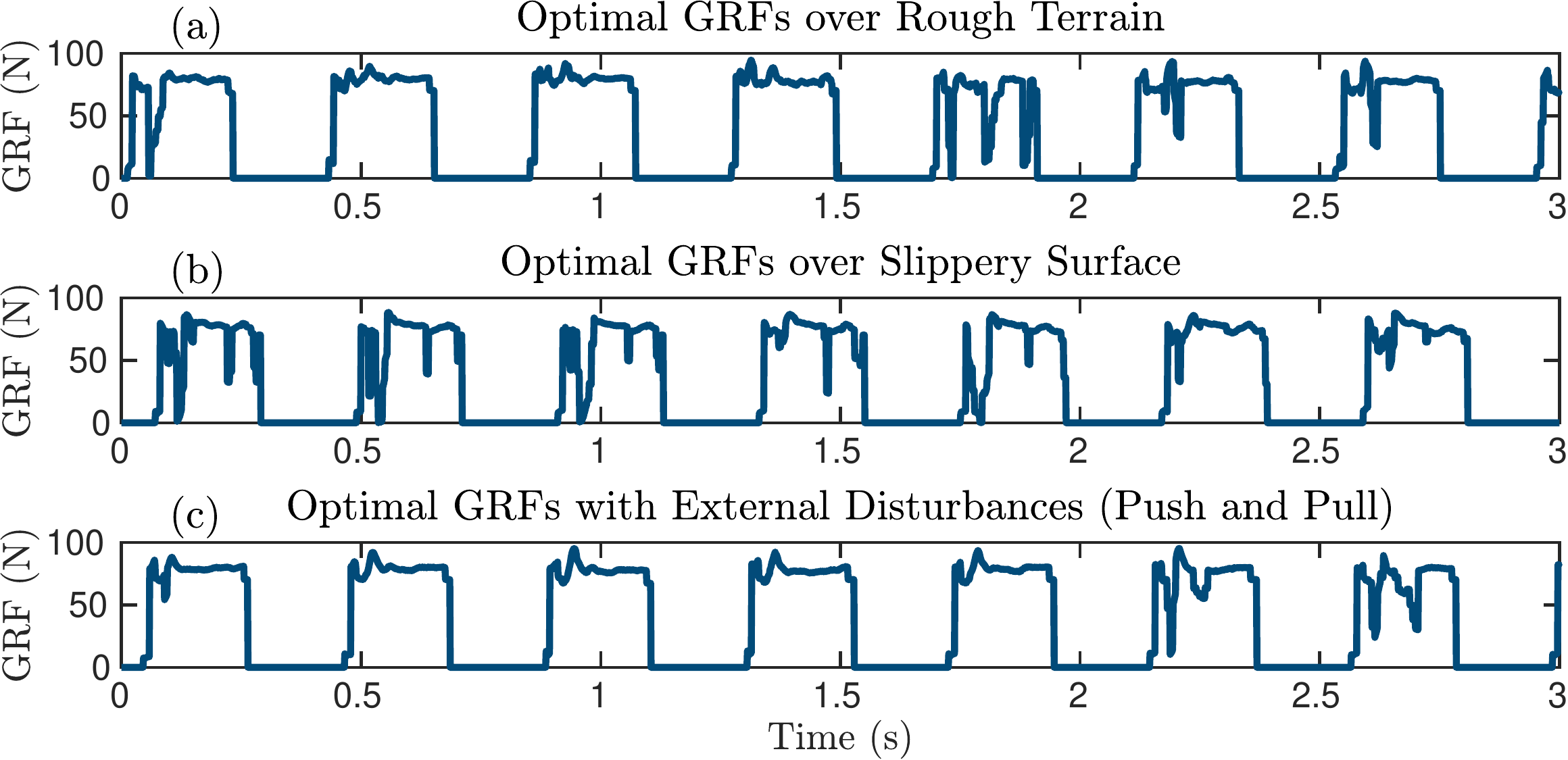}
\vspace{-1.5em}
\caption{Plots of the optimal GRFs, generated by the centralized MPC, for cooperative locomotion experiments on (a) rough terrain, (b) a slippery surface, and (c) subject to external disturbances. The figure depicts the optimal GRFs for the left front leg of agent 1 along the $z$-direction.}
\label{fig:grf_disturb}
\vspace{-0.5em}
\end{figure}

\begin{figure}[t!]
\centering
\includegraphics[draft=false, width=\linewidth]{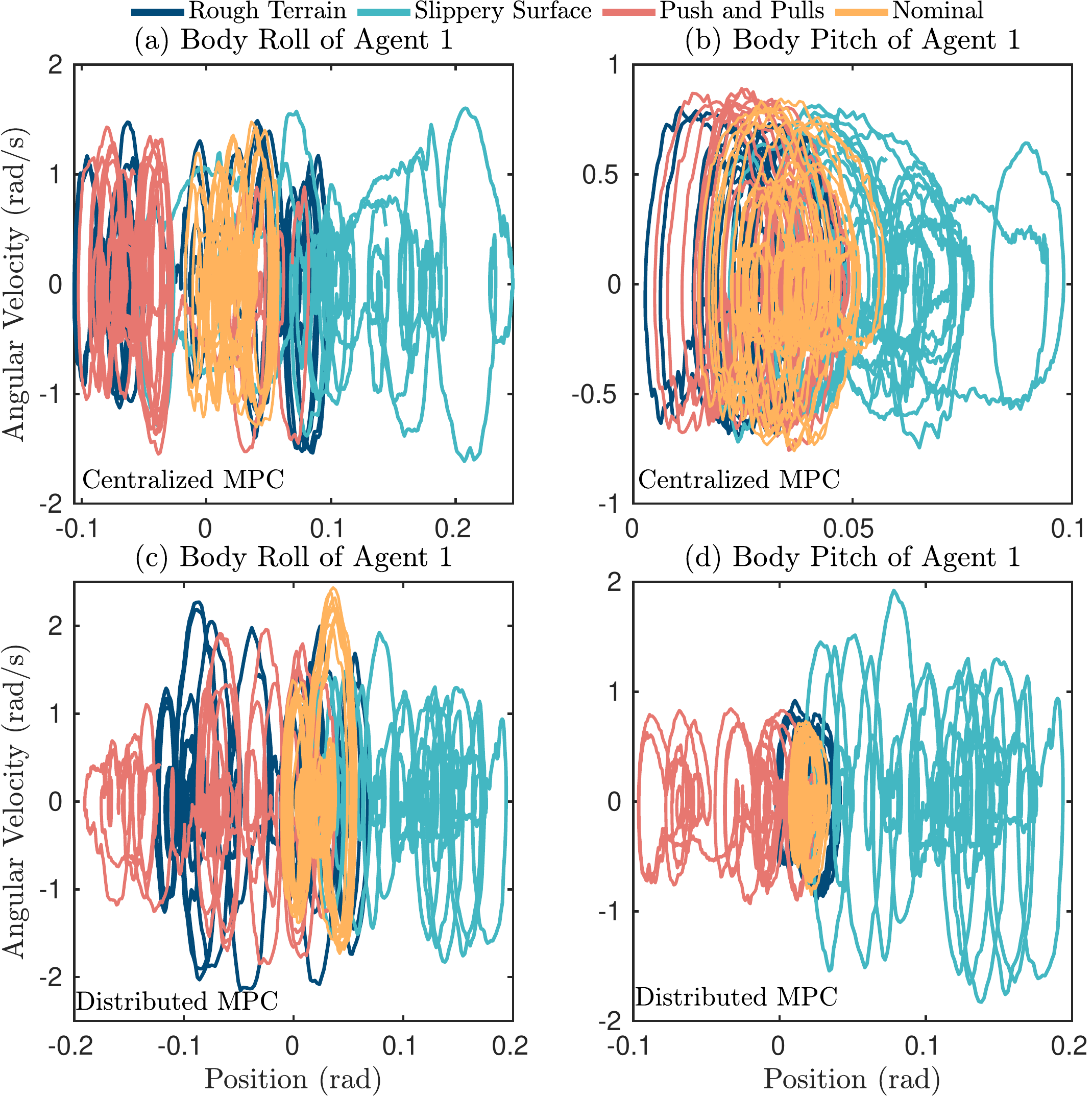}
\vspace{-1.5em}
\caption{Phase portraits for (a) the body roll and (b) the body pitch of agent $1$ with the centralized MPC and (c) the body roll and (d) the body pitch of agent $1$ with the distributed MPCs during different experiments. The plots show the robustness of the cooperative locomotion over rough terrain covered with randomly dispersed wooden blocks, the slippery surface, and subject to unknown external disturbances.}
\label{fig:phaseportraits_c_and_d}
\vspace{-0.5em}
\end{figure}

We further investigate the robustness of the proposed layered control approach by studying the tracking performance of the closed-loop system with different experiments, including locomotion on rough terrain (see Fig. \ref{fig:indoor_exp}(a)), locomotion on a slippery surface (see Fig. \ref{fig:indoor_exp}(c)), and locomotion subject to unknown external disturbances (see Fig. \ref{fig:indoor_exp}(d)), as shown in Figs. \ref{fig:trot_vel_centralized_disturb}(a), \ref{fig:trot_vel_centralized_disturb}(b), and \ref{fig:trot_vel_centralized_disturb}(c), respectively. In these experiments, the rough terrain is made of randomly displaced wooden blocks with a maximum height of 5 (cm) ($19\%$ of the robot's height). Moreover, the slippery surface is a whiteboard covered with cooking spray. The unknown external disturbances are further applied by a human user, including pushes and tethered pulls on both agents. The robots cooperatively transport an unknown payload of 4.53 (kg) ($36\%$ uncertainty in one robot's mass) in all these experiments. The optimal GRFs computed by the MPC on rough terrain, on the slippery surface, and subject to external disturbances are depicted in Figs. \ref{fig:grf_disturb}(a), \ref{fig:grf_disturb}(b), and \ref{fig:grf_disturb}(c), respectively. We remark that despite the uncertainties, the GRFs are in the feasible range, and the MPC's outputs robustly track the desired and time-varying trajectories. Furthermore, the phase portraits of the body's roll and pitch motions (i.e., unactuated DOFs) during these cooperative trot gaits are shown in Figs. \ref{fig:phaseportraits_c_and_d}(a) and \ref{fig:phaseportraits_c_and_d}(b). Figure \ref{fig:phaseportraits_c_and_d} indicates that the A1 robots can perform robustly stable cooperative locomotion in the presence of various unknown terrains and disturbances. Videos of all experiments are available online \cite{YouTube_CoopertaiveLoco}.

\begin{figure}[t!]
\centering
\includegraphics[draft=false, width=\linewidth]{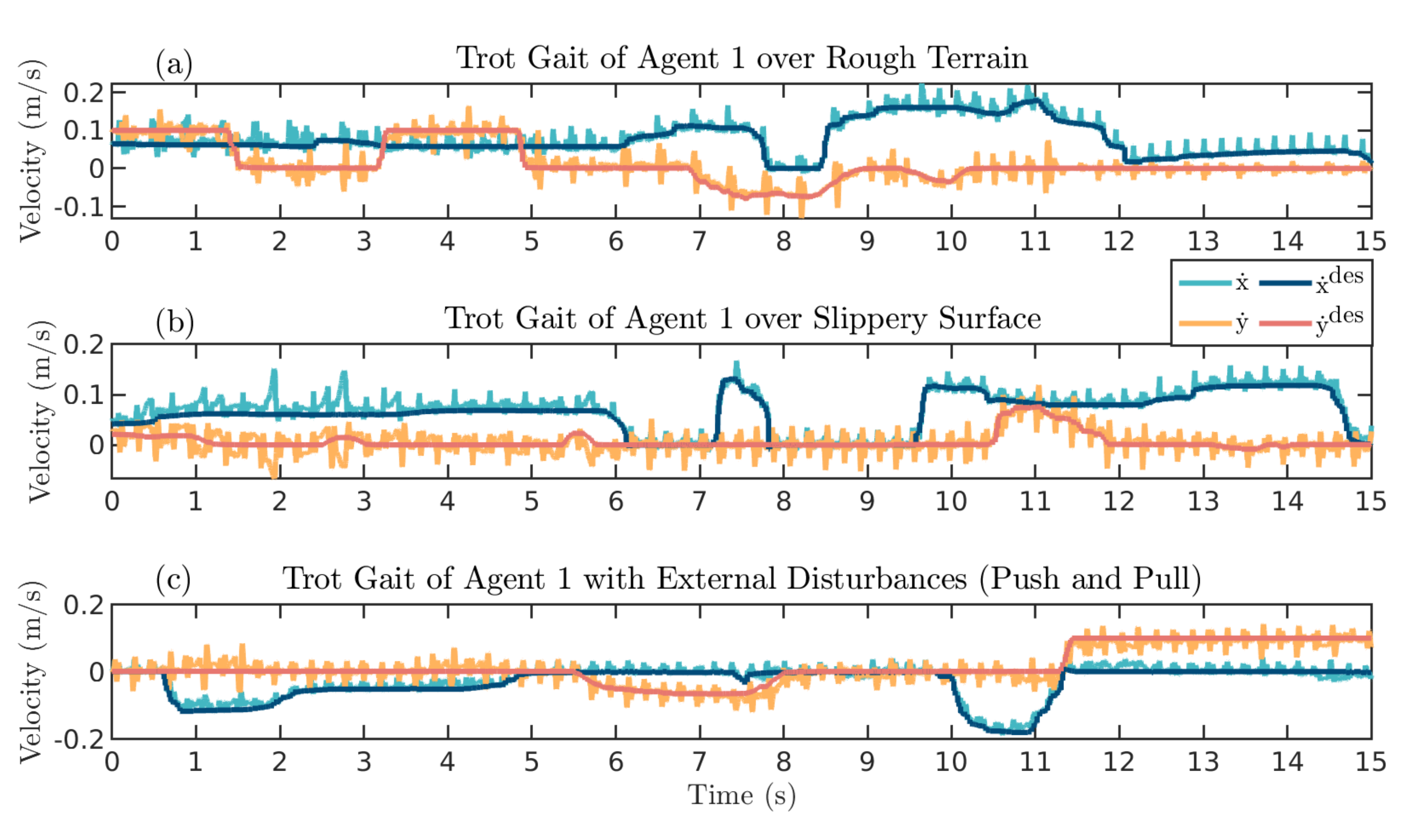}
\vspace{-1.5em}
\caption{Plots of the desired and optimal velocities for cooperative locomotion experiments on (a) rough terrain, (b) a slippery surface, and (c) subject to external disturbances with the distributed MPC. Time-varying desired trajectories are provided by the joystick. It is clear that the distributed MPC's outputs can robustly track the desired trajectories in the presence of uncertainties and disturbances.}
\label{fig:trot_vel_disturb_distributed}
\vspace{-0.5em}
\end{figure}

\begin{figure}[t!]
\centering
\includegraphics[draft=false, width=\linewidth]{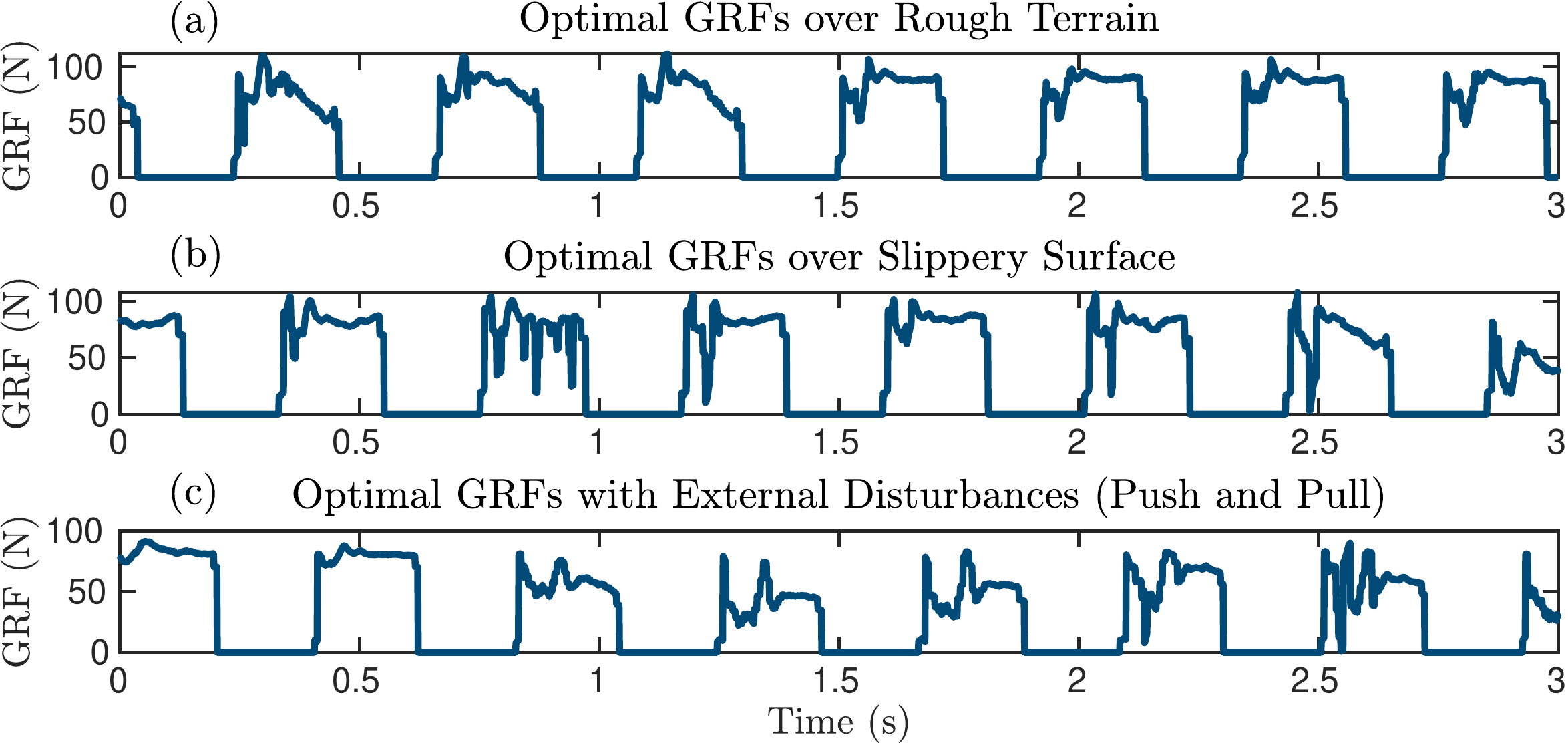}
\vspace{-1.5em}
\caption{Plots of the optimal GRFs, generated by the distributed MPC, for cooperative locomotion experiments on (a) rough terrain, (b) a slippery surface, and (c) subject to external disturbances. The figure depicts the optimal GRFs for the left front leg of agent 1 along the $z$-direction.}
\label{fig:grf_disturb_distributed}
\vspace{-0.5em}
\end{figure}

\subsubsection{\textbf{Indoor experiments with the distributed MPC}}
In this part, we evaluate the performance of the closed-loop system with the proposed distributed MPC algorithm in similar indoor experiments (see Fig. \ref{fig:indoor_exp}).  The evolution of the optimal trajectories generated from the distributed MPC and time-varying desired trajectories during the cooperative transportation of the same payload over rough terrain, the slippery surface, and subject to unknown disturbances are illustrated in Figs. \ref{fig:trot_vel_disturb_distributed}(a), \ref{fig:trot_vel_disturb_distributed}(b), and \ref{fig:trot_vel_disturb_distributed}(c), respectively. The optimal GRFs are also shown in Fig. \ref{fig:grf_disturb_distributed}. The phase portraits of the body's roll and pitch motions during the cooperative gait with the distributed MPC algorithm and subject to these uncertainties are depicted in Figs. \ref{fig:phaseportraits_c_and_d}(c) and \ref{fig:phaseportraits_c_and_d}(d). It is evident that the optimal GRFs, generated by the MPC, remain feasible, and the MPC's outputs robustly track the desired trajectories in the presence of unknown terrains and external disturbances.

\begin{figure}[t!]
\centering
\includegraphics[width=\linewidth]{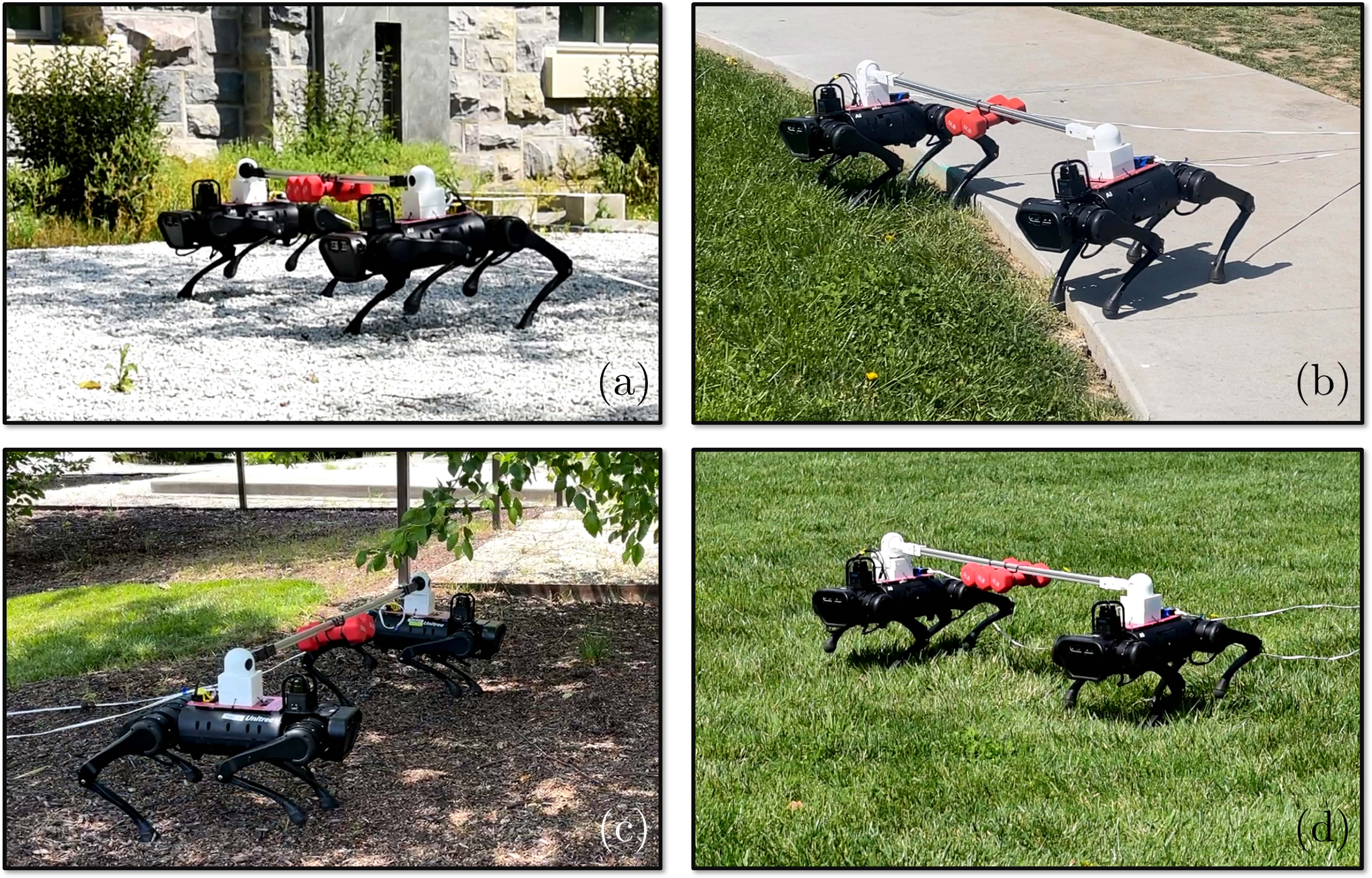}
\vspace{-1.5em}
\caption{Snapshots demonstrate the proposed layered controller's performance for a series of cooperative locomotion experiments. Outdoor experiments: (a) cooperative locomotion on gravel, (b) transitioning from concrete surface to grass, (c) cooperative locomotion on mulch, and (d) cooperative locomotion on grass. The robots cooperatively transport a payload of 4.53 (kg) ($36\%$ uncertainty) in (b) and (c) and 6.80 (kg) ($55\%$ uncertainty) in (a) and (d). Here, (a) and (c) show the snapshots where the distributed MPC is adopted, while (b) and (d) show the snapshots where the centralized MPC is employed.}
\vspace{-0.5em}
\label{fig:outdoor_exp}
\end{figure}

\begin{figure}[t!]
\centering
\includegraphics[draft=false, width=\linewidth]{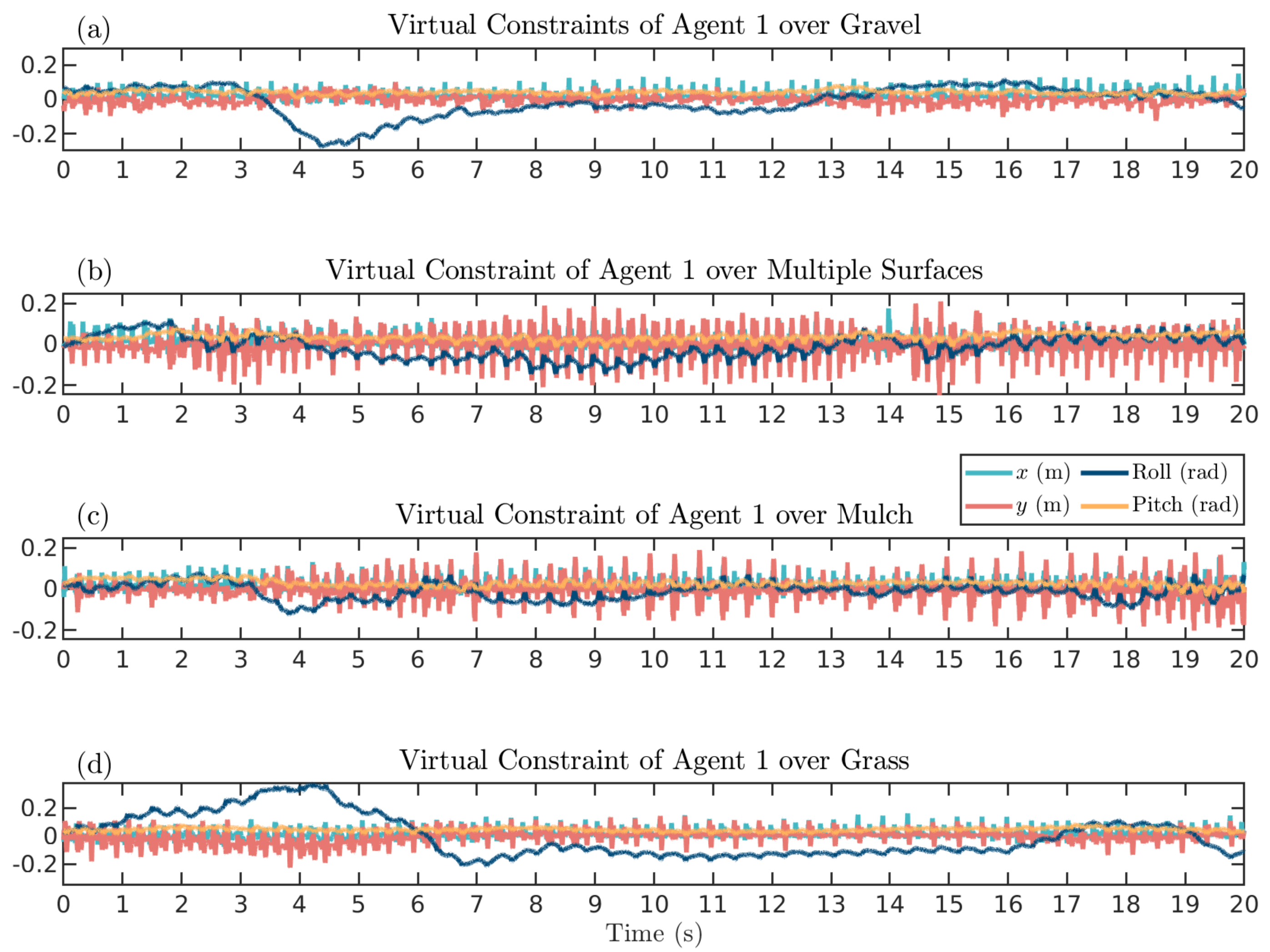}
\vspace{-1.5em}
\caption{Plots of the virtual constraints of agent 1 during cooperative locomotion with unknown payloads and on various outdoor terrains, including (a) locomotion on gravel, (b) transitioning from concrete to grass, (c) locomotion on mulch, and (d) locomotion on grass. The payload is 4.53 (kg) in (b) and (c) and 6.80 (kg) in (a) and (d). Here, (a) and (c) depict the evolution of virtual constraints with the distributed MPC at the high level. In addition, (b) and (d) illustrate the evolution of the virtual constraints with the centralized MPC at the high level. Here, we plot the components of virtual constraints in \eqref{virtual_constraints} that correspond to the COM position along the $x$ and $y$ axes (m) (i.e., COM position tracking) and the body’s roll and pitch angles (rad) (i.e., angle tracking). It is clear that the full-order system tracks the prescribed optimal and reduced-order trajectories generated by the MPCs.}
\label{fig:outdoor_vc}
\vspace{-0.5em}
\end{figure}

\subsubsection{\textbf{Outdoor experiments with centralized and distributed MPCs}}
We next investigate the performance and robustness of the closed-loop system with the centralized and distributed MPC algorithms in different outdoor experiments, as shown in Fig. \ref{fig:outdoor_exp}. These experiments include cooperative locomotion on gravel, concrete, mulch, and grass subject to unknown payloads. In these studies, we investigate two different payloads: a payload of 4.53 (kg) ($36\%$ uncertainty) in Figs. \ref{fig:outdoor_exp}(b) and \ref{fig:outdoor_exp}(c) and a payload of 6.80 (kg) ($55\%$ uncertainty) in Figs. \ref{fig:outdoor_exp}(a) and \ref{fig:outdoor_exp}(d). The evolution of the virtual constraints \eqref{virtual_constraints} for trotting over the gravel and transitioning from concrete to grass with the centralized MPC and trotting over mulch and grass with the distributed MPC is shown in Fig. \ref{fig:outdoor_vc}. As the virtual constraint plots stay close to zero, we can conclude that the full-order system effectively tracks the optimal reduced-order trajectories generated by the high-level MPCs. Furthermore, it is evident that the proposed layered control approach with both centralized and distributed MPCs can robustly stabilize cooperative gaits in the presence of payloads on unknown outdoor terrains. 


\section{Discussion and Comparison}
\label{Sec:Discussion and Comparison}

Numerical simulations and experimental validations in Section \ref{Sec:Numerical and Experimental Validations} show the effectiveness of the proposed centralized and distributed MPC algorithms for cooperative locomotion. This section aims to analyze and compare the performance of the proposed MPCs while discussing their limitations.

\begin{figure*}[t!]
\centering
\includegraphics[draft=false, width=\linewidth]{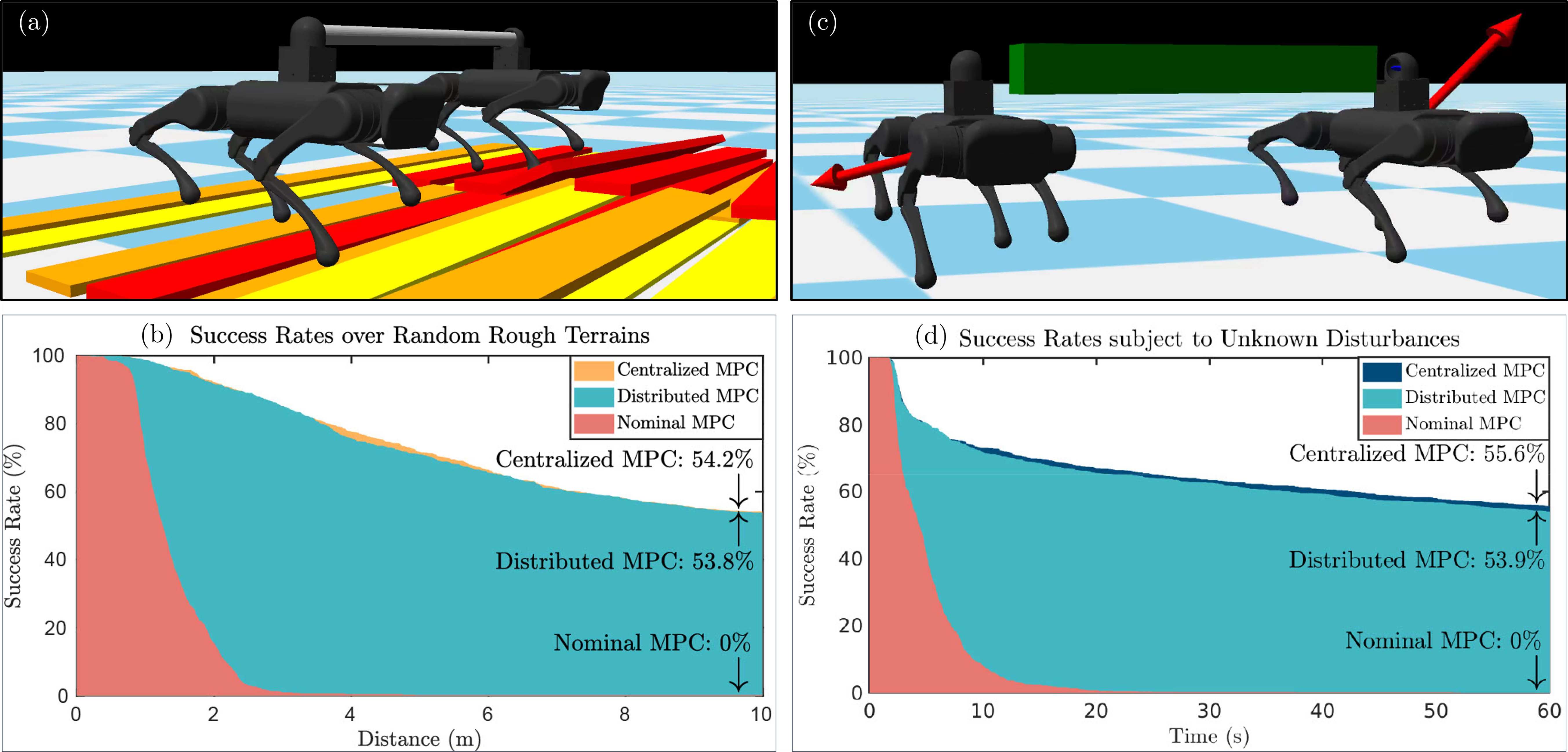}
\vspace{-1.5em}

\caption{Illustration of the comparison results between the nominal, centralized, and distributed MPCs. (a) The snapshot shows the RaiSim simulation environment with one of the randomly generated rough terrains. The maximum height of the generated terrains is 12 (cm) ($46\%$ uncertainty in robots' height). (b) The plot describes the success rate of the proposed trajectory planners over 1500 randomly generated rough terrains in numerical simulations. The overall success rate of the nominal, centralized, and distributed MPCs over randomly generated rough terrain is $0\%$, $54.2\%$, and $53.8\%$, respectively. (c) The snapshot shows the RaiSim simulation environment with one of the randomly generated external forces and a randomly generated payload. The arrows illustrate the applied external forces on each agent. The maximum external force is 80 (N) ($65\%$ of one robot's weight) on the $x$-, $y$-, and $z$-directions. The evolution of the forces in each direction is sinusoidal with a maximum random period of 4 (s). External forces are applied from 1 (s) to 60 (s). The maximum payload mass is 5 (kg).  (d) The plot describes the success rate of the proposed trajectory planners with 1200 randomly applied external forces and payloads in numerical simulations. The overall success rate of the nominal, centralized, and distributed MPCs subject to unknown external forces and payloads is $0\%$, $55.6\%$, and $53.9\%$, respectively.}
\label{fig:comp_total}
\end{figure*}

\vspace{-0.5em}
\subsection{Comparison of the Centralized and Distributed MPCs}
\label{subsec:comp_c_d}

The robustness of the cooperative locomotion with the proposed centralized and distributed MPC algorithms in the presence of various uncertainties and disturbances is studied numerically and experimentally in Section \ref{Sec:Numerical and Experimental Validations}. To compare the performance and robustness of the proposed trajectory planners, we apply the nominal, centralized, and distributed MPCs over 1500 randomly generated rough terrains in the simulation environment of RaiSim, as shown in Fig. \ref{fig:comp_total}(a). Here, the randomly generated landscapes' maximum height is 12 (cm) ($46\%$ uncertainty in the robot's height). Furthermore, the total length of the terrain is assumed to be 10 (m). In these simulations, we evaluate the cooperative locomotion as a success if the agents reach 10 (m) without losing stability. We assess the locomotion as a failure if at least one of the agents' bodies touches the ground before reaching 10 (m). The success rate versus the length of the terrain is depicted in Fig. \ref{fig:comp_total}(b). The overall success rate of the nominal, centralized, and distributed MPCs is $0\%$, $54.2\%$, and $53.8\%$, respectively.

Similarly, we compare the performance and robustness of the nominal, centralized, and distributed MPCs subject to 1200 randomly generated external forces and payloads, as shown in Fig. \ref{fig:comp_total}(c). The external force is taken as sinusoidal with a maximum amplitude of 80 (N) ($65\%$ of one robot's weight) and a maximum period of 4 (s) on the $x$-, $y$-, and $z$-directions. The maximum mass of the payload is also assumed to be 5 (kg). We evaluate the cooperative locomotion as a success if the agents sustain the stability until 60 seconds. We assess the locomotion as a failure if at least one of the agents' bodies touches the ground before 60 (s). The success rate versus time is depicted in Fig. \ref{fig:comp_total}(d). The overall success rate of the nominal, centralized, and distributed MPCs is $0\%$, $55.6\%$, and $53.9\%$, respectively.

Our experimental studies in Figs. \ref{fig:forwardtrot_centralized}-\ref{fig:grf_disturb_distributed} and Fig. \ref{fig:outdoor_vc} suggest that the proposed centralized and distributed trajectory planners show similar robustness in indoor and outdoor experiments. Slightly better robustness has been observed in numerical simulations of Fig. \ref{fig:comp_total} when employing the centralized MPC at the high level. Still, the success rate between the centralized and distributed MPCs does not significantly differ. These comparisons suggest that the proposed centralized and distributed MPCs can robustly stabilize dynamic cooperative locomotion. However, the distributed MPC has substantially less computational time.

\begin{figure}[t!]
\centering
\includegraphics[draft=false, width=\linewidth]{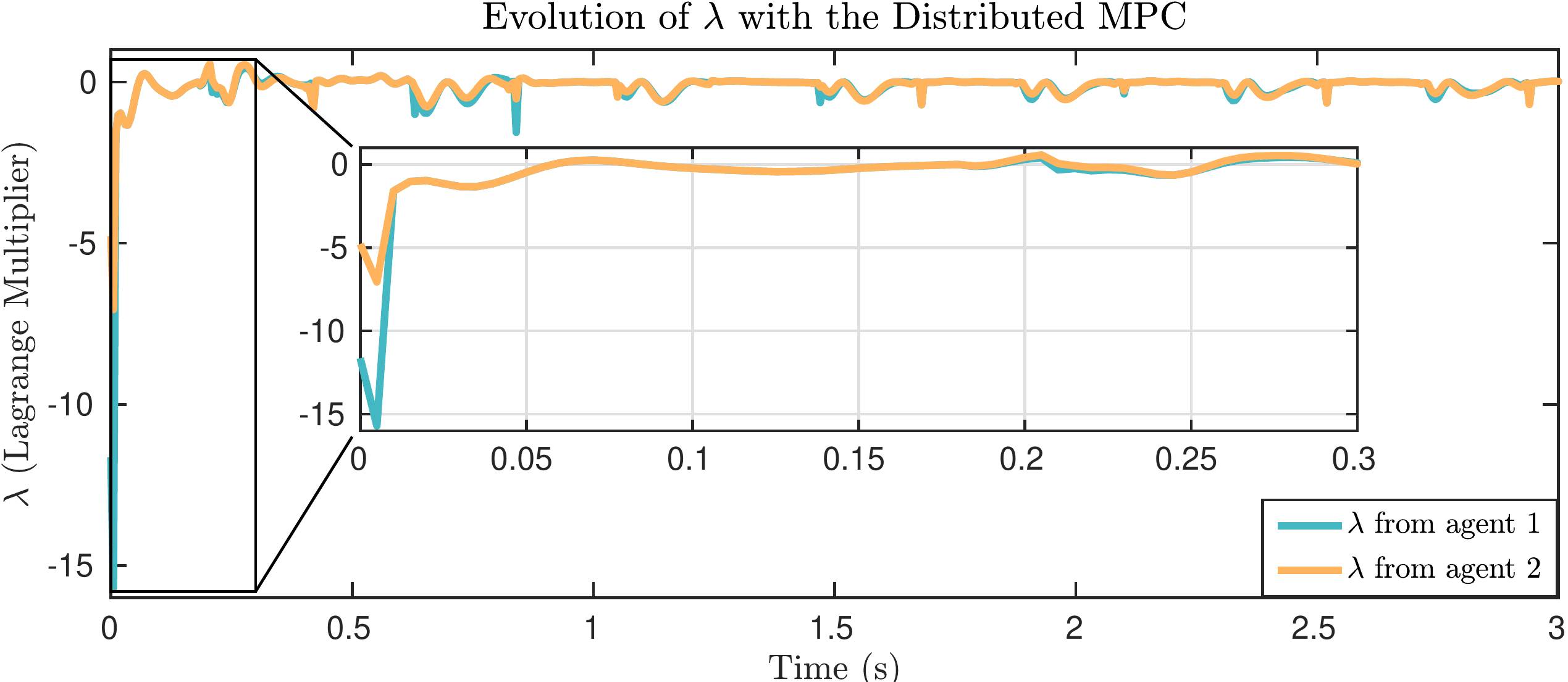}
\vspace{-1.5em}
\caption{Plots of the locally estimated Lagrange multiplier $\lambda$ by two agents using the distributed MPC algorithm. It is clear that the local values reach an agreement and stay close to each other.}
\label{fig:lambdaevolution}
\vspace{-0.5em}
\end{figure}

\vspace{-0.5em}
\subsection{Evolution of the Lagrange Multiplier in Distributed MPC}
\label{subsec:lambdaevolution}

Section \ref{subsec:comp_c_d} demonstrated a similar success rate for the centralized and distributed MPC algorithms with the randomly generated terrains and disturbances. To further study this similar robust stability behavior, Fig. \ref{fig:lambdaevolution} illustrates the evolution of the estimated Lagrange multiplier, $\lambda$, for each agent when the agents cooperatively walk with the distributed MPC. In formulating the distributed MPC, each agent locally estimates the Lagrange multiplier according to the one-step communication delay and the agreement protocol. Therefore, $\lambda$ on each distributed MPC evolves differently. We introduced the consensus protocol in the cost function of \eqref{distributed_MPC} to mitigate the divergence of the local estimates and to impose the agreement. The magnified portion of the plot in Fig. \ref{fig:lambdaevolution} shows that the initial $\lambda$ values on each agent are different while converging after a short amount of time according to the consensus protocol. The plot also shows that each agent's $\lambda$ values are not precisely the same during cooperative locomotion. However, we can observe that both $\lambda$ values stay close.

\begin{figure}[t!]
\centering
\includegraphics[draft=false, width=\linewidth]{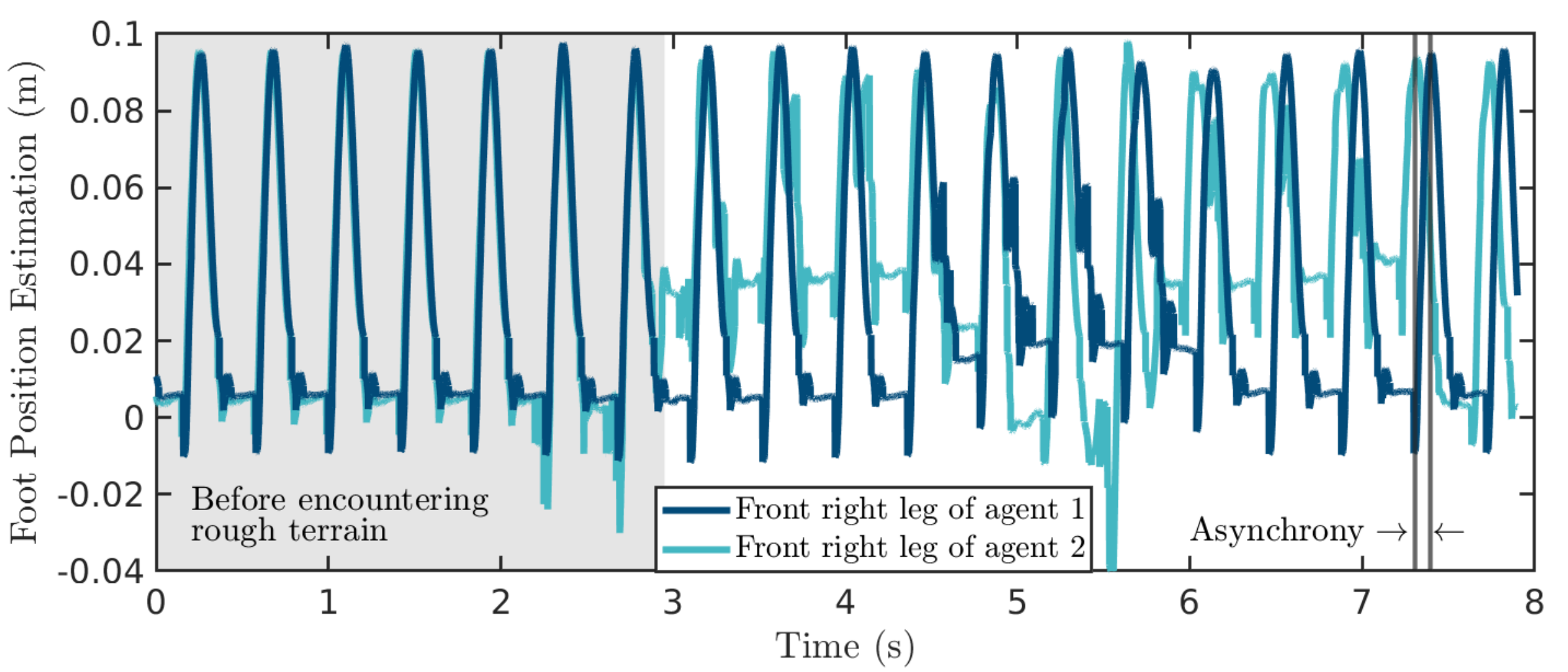}
\vspace{-1.5em}
\caption{Plots of the estimated height of agents' front right legs. Initial locomotion has complete synchrony before encountering the rough terrain described in the gray area in the plot. After engaging the rough terrain, it is clear that asynchrony happens between agents. However, the layered control approach robustly stabilizes the cooperative gait.}
\label{fig:asynchrony}
\vspace{-0.5em}
\end{figure}

\vspace{-0.5em}
\subsection{Synchronization and Asynchronization}
\label{subsec:syncvsasync}

We aim to study the robustness of the layered control approach against possible phase differences between agents that can easily occur on rough terrain, where the discrete-time transitions (i.e., impacts) happen earlier or later than anticipated times on normal gaits. To further investigate this point, we study the estimated height of the agents' front right legs over rough terrain in Fig. \ref{fig:asynchrony}. Both agents are synchronized at the beginning of the locomotion. After encountering the rough terrain, the asynchrony is observed in Fig. \ref{fig:asynchrony}. However, the proposed centralized and distributed MPCs show robust cooperative gaits over unknown rough terrains, as shown in Figs. \ref{fig:indoor_exp}(a), \ref{fig:phaseportraits_c_and_d}, \ref{fig:outdoor_exp}, and Fig. \ref{fig:outdoor_vc}. Moreover, the robustness subject to more than 1000 randomly generated rough terrains is also validated in Fig. \ref{fig:comp_total}.

\begin{figure}[t!]
\centering
\includegraphics[draft=false, width=\linewidth]{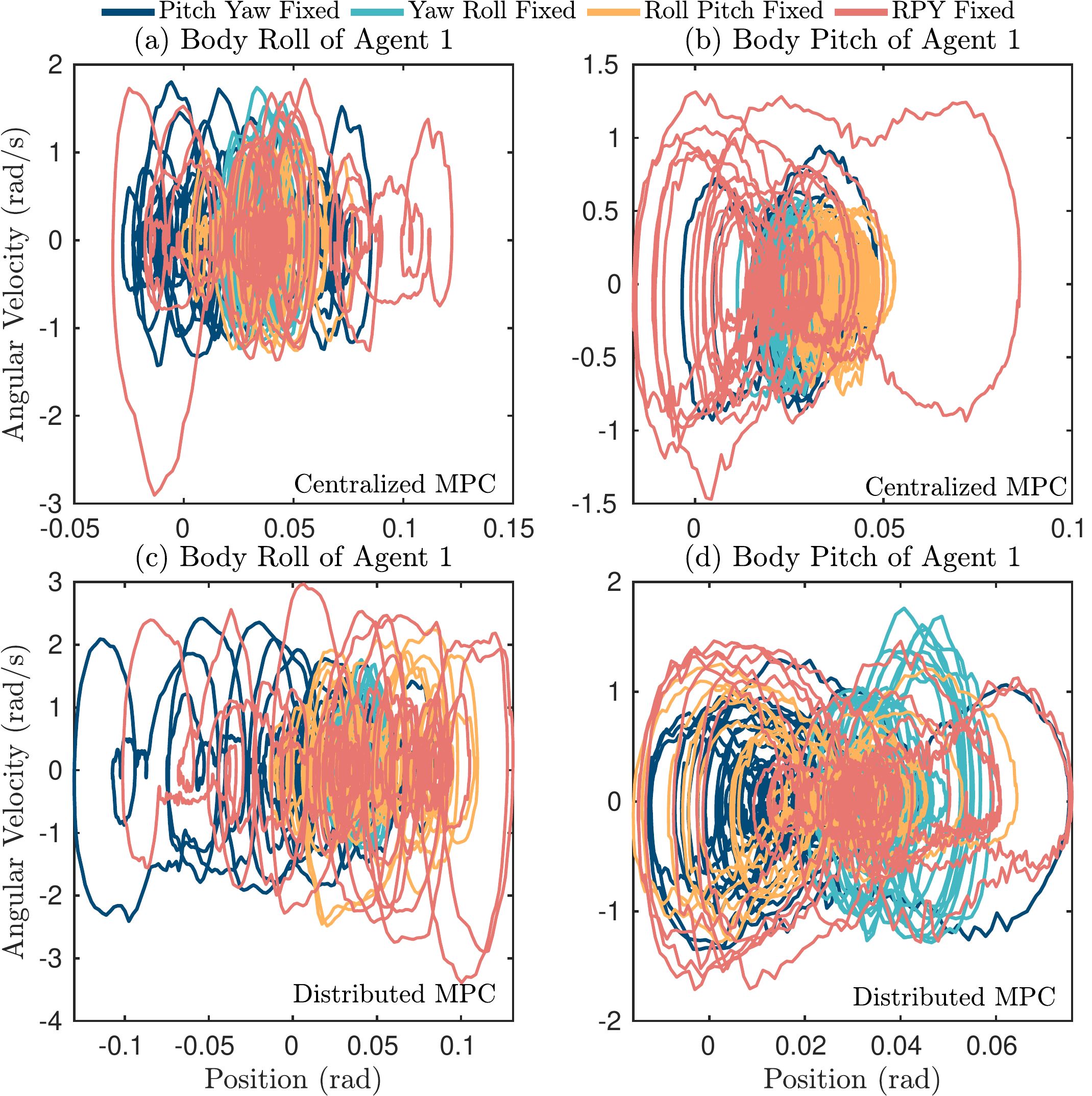}
\vspace{-1.5em}
\caption{Phase portraits for (a) the body roll and (b) the body pitch of agent $1$ with the centralized MPC and (c) the body roll and (d) the body pitch of agent $1$ with the distributed MPC during different experiments. The plots show the robustness of the cooperative locomotion over rough terrain with fixed DOFs in holonomic constraints on the roll, pitch, and yaw directions.}
\label{fig:phaseportraits_dof_reduce}
\vspace{-0.5em}
\end{figure}

\vspace{-0.5em}
\subsection{Robustness Against Unknown Holonomic Constraints}
\label{subsec:rotationregulation}

The holonomic constraint of Section \ref{Sec:Reduced-Order Model of Cooperative Locomotion} assumes a distance constraint between the interaction points of agents. In particular, we take no additional rotational constraints at the interaction points. This assumption simplifies the interconnected SRB model and, thereby, the centralized and distributed MPC algorithms. However, more sophisticated connections could exist, such as limited DOFs on both ends of the holonomic constraint. Here, we study the robustness of the proposed MPCs subject to uncertainties arising from rotational restrictions at the interaction points. These constraints can arise from cooperative loco-manipulation in various applications. 
Figure \ref{fig:phaseportraits_dof_reduce} depicts the body roll and pitch evolution during cooperative locomotion over rough terrain with different holonomic constraints at the interaction points, including restrictions on ball joints' pitch-yaw, yaw-roll, roll-pitch, and roll-pitch-yaw. These restrictions are implemented with the different mechanisms designed in Fig \ref{fig:dofregulation}. The robust stability of the cooperative locomotion with the proposed centralized MPC is shown in the phase portraits of the body roll and body pitch in Figs. \ref{fig:phaseportraits_dof_reduce}(a) and \ref{fig:phaseportraits_dof_reduce}(b). The robust stability of the proposed distributed MPC is also illustrated in Figs. \ref{fig:phaseportraits_dof_reduce}(c) and \ref{fig:phaseportraits_dof_reduce}(d). We observe that the cooperative locomotion over rough terrain with different and unknown holonomic constraints has robust stability similar to the one illustrated in Fig. \ref{fig:phaseportraits_c_and_d}. However, the phase portraits in Fig. \ref{fig:phaseportraits_dof_reduce} show that the unknown additional interactions from the limited DOFs on both ends of the holonomic constraint can induce aggressive angular positions and velocity changes.


\vspace{-0.5em}
\subsection{Limitations and Future Study}
\label{subsec:Limitations and Future Study}

\subsubsection{\textbf{Optimal control with switching}}

The proposed MPC approaches for cooperative locomotion were shown to be very robust to various unknown terrains and subject to unknown disturbances. However, the gait presented here does not exhibit extremely dynamic or highly agile maneuvers. One of the reasons for this is the relatively small planning horizon ($25$ (ms)). While the distributed approach provides an interesting avenue to explore longer horizons in future work due to the considerable decrease in computation time, long horizons suffer when only considering the current domain. For this reason, future work should not only explore increased planning lengths but should also consider a PWA optimal control formulation \cite[Chap. 16]{MPC_Book} such that the change in stance leg configurations can be considered directly by the planner.

\subsubsection{\textbf{Sophisticated constraints between agents}}
We assumed the holonomic constraint \eqref{holonomic_constraint} with a ball joint on agents to simplify the development of the interconnected reduced-order model and the synthesis of centralized and distributed MPCs. We further studied the robustness of the proposed layered control approach subject to the unknown restrictions on the ball joints in Section \ref{subsec:rotationregulation}. However, more sophisticated cooperative tasks may require dexterous manipulation during cooperative locomotion. For instance, quadrupedal robots can be equipped with robotic arms for loco-manipulation. Our future work will investigate the development of robust control algorithms that systematically address the gap between simplified reduced-order models and complex dynamical models of cooperative loco-manipulation.

\subsubsection{\textbf{Extension to multi-agents}} Our previous work \cite{Jeeseop_Hamed_ASME} presented quasi-statically stable cooperative gaits for $M\geq2$ agents. In particular, a closed-form expression for the interconnected LIP models was developed to address the real-time trajectory planning based on a centralized MPC algorithm. The interconnected LIP model \textit{cannot} address interaction torques between the agents. Furthermore, the gait is \textit{not} dynamic. The current paper presents an interconnected reduced-order model, based on the SRB dynamics, that addresses interaction torques between the agents while allowing dynamic cooperative gaits. In addition, centralized and distributed MPC algorithms are developed for the cooperative locomotion of two agents. However, a closed-form expression for the Jacobian matrices in \eqref{prediction_dyn} and \eqref{equality_constrinats} may not be easily computed for $M\geq3$ interconnected SRB dynamics with sophisticated holonomic constraints. Our future work will investigate the extension of the approach for dynamic cooperative locomotion of $M\geq3$ agents with complex holonomic constraints. One possible way is to develop robust distributed MPC algorithms integrated with reinforcement learning and data-driven techniques \cite{pandala2022robust,fawcett2022toward} to bridge the gap between interconnected reduced-order models and full-order models. 

\subsubsection{\textbf{Coordination between agents}}
In numerical simulations, each agent's global coordinates can be easily used without sensor limitations or unexpected noises. However, experimental evaluations estimate the agents' global coordinates via kinematic estimators. The estimation errors may result in unexpected coordination changes. This paper addresses this issue by the human operator who coordinates the agents with the corresponding speed commands from the joystick. For example, the user commands a higher or lower desired speed to the lagging or leading agent, respectively. Our future work will investigate the design of algorithms that robustly estimate the global coordinates of the agents in the presence of noisy measurements. 


\section{Conclusion}
\label{Sec:Conclusion and Future Work}

This paper presented a layered control algorithm for real-time trajectory planning and robust control for cooperative locomotion of two holonomically constrained quadrupedal robots. An innovative reduced-order model of cooperative locomotion is developed and studied based on interconnected SRB dynamics. At the high level of the layered control algorithm, the real-time trajectory planning problem is formulated as an optimal control problem of the interconnected reduced-order model with two different schemes: centralized and distributed MPCs. The centralized MPC plans for the global reduced-order states, global GRFs, and the interaction wrenches between agents. The distributed MPC is developed based on a one-step communication delay and an agreement protocol to solve for the local reduced-order states, local GRFs, and the local estimated wrenches. At the low level of the control scheme, distributed nonlinear controllers, based on QP and virtual constraints, are developed to impose the full-order model of each agent to track the optimal reduced-order trajectories and GRFs prescribed by the high-level MPCs.

The effectiveness of the proposed layered control approach was verified with extensive numerical simulations and experiments for the blind and robust cooperative locomotion of two holonomically constrained A1 robots with different payloads on different terrains and subject to external disturbances. A detailed study was presented to compare the performance of the proposed centralized and distributed MPCs over more than 1000 randomly generated landscapes and external pushes. It was shown that the distributed MPC has a robust stability performance similar to that of the centralized MPC, while the computation time is reduced significantly. The results also show that both the centralized and distributed MPCs integrated with the interconnected SRB dynamics can drastically improve the robust stability of cooperative locomotion compared to the individual nominal MPCs. The experimental results suggest that the proposed control algorithm can result in robustly stable cooperative locomotion on different terrains (e.g., wooden blocks, slippery surfaces, grass, mulch, and concrete) subject to unknown payloads and external disturbances at different speeds. The robustness of the control approach was also studied against uncertainties in holonomic constraints and assumptions. 

For future work, we will investigate the extension of the approach for more sophisticated constraints between agents. We will also study the extension to multi-agents while systematically developing robust optimal control algorithms to address switching in hybrid models.  


\bibliographystyle{IEEEtran}
\bibliography{references}
\end{document}